\documentclass{article}
\usepackage[table]{xcolor}% http://ctan.org/pkg/xcolor
\usepackage{microtype}
\usepackage{graphicx}
\usepackage{subfigure}
\usepackage{booktabs} % for professional tables
\usepackage{makecell}
\usepackage{multirow}
\usepackage{tabularx}

\usepackage{hyperref}

\usepackage[accepted]{icml2024}

\usepackage{amsmath}
\usepackage{amssymb}
\usepackage{mathtools}
\usepackage{amsthm}

\usepackage[capitalize,noabbrev]{cleveref}

\theoremstyle{plain}

\theoremstyle{definition}

\theoremstyle{remark}

\usepackage[textsize=tiny]{todonotes}

\usepackage{ulem}

\icmltitlerunning{Latent Logic Tree Extraction for Event Sequence Explanation from LLMs}

\begin{document}

\twocolumn[
\icmltitle{
% Amortizing Structural Learning for Event Sequences in Large Language Model
Latent Logic Tree Extraction for Event Sequence Explanation from LLMs
}




\begin{icmlauthorlist}
\icmlauthor{Zitao Song}{comp}
\icmlauthor{Chao Yang}{yyy}
\icmlauthor{Chaojie Wang}{sch}
\icmlauthor{Bo An}{sch,comp}
\icmlauthor{Shuang Li}{yyy}
\end{icmlauthorlist}

\icmlaffiliation{comp}{School of Computer Science and Engineering, Nanyang Technological University, Singapore}
\icmlaffiliation{yyy}{School of Data Science, The Chinese University of Hong Kong, Shenzhen, China}
\icmlaffiliation{sch}{Skywork AI, Singapore}

\icmlcorrespondingauthor{Shuang Li}{lishuang@cuhk.edu.cn}

\icmlkeywords{Machine Learning, ICML}

\vskip 0.3in
]



\printAffiliationsAndNotice{}  % leave blank if no need to mention equal contribution

\begin{abstract}
%   Contemporary data collection across fields like social media, healthcare, and robotics generates vast streaming event sequences characterized by complex latent temporal patterns and social behaviors. Traditional neural network-based models often struggle to accurately capture these patterns, leading to subpar predictive outcomes. This challenge has inspired the use of Large Language Models (LLMs), which leverage their pretrained prior knowledge to learn structural and semantic cross-event dependencies for improved future inference. In this work, we interpret tree structure \bb{(tree-structured)} learning for event sequence as a Latent Variable Modeling (LVM) problem. We present \textbf{LTS-TPP}, an EM-style framework for learning \textbf{L}atent \textbf{T}ree-\textbf{S}tructure underlying the \textbf{T}emporal \textbf{P}oint \textbf{P}rocess, where the E-step involves employing LLMs to sample a latent symbolic tree structure, with posterior approximation achieved by finetuning LLMs using diversity-seeking reinforcement learning algorithms.
% % We amortize the intractable posterior of the symbolic tree by finetuning LLMs via diversity-seeking reinforcement learning algorithms: generative flow networks (GFlowNets). 
% In the M-step, given the generated symbolic tree structure, the framework will maximize a \textit{modular} likelihood of the event history. Empirical evidence shows that our approach of aligning LLM fine-tuning with event sequences results in unparalleled performance in both analyzing and predicting real-world user behavior datasets.

Modern high-stakes systems, such as healthcare or robotics, often generate vast streaming event sequences. Our goal is to design an efficient, plug-and-play tool to elicit logic tree-based explanations from Large Language Models (LLMs) to provide customized insights into each observed event sequence. Built on the temporal point process model for events, our method employs the likelihood function as a score to evaluate generated logic trees. We propose an amortized Expectation-Maximization (EM) learning framework and treat the logic tree as latent variables. In the E-step, we evaluate the posterior distribution over the latent logic trees using an LLM prior and the likelihood of the observed event sequences. LLM provides a high-quality prior for the latent logic trees, however, since the posterior is built over a discrete combinatorial space, we cannot get the closed-form solution. We propose to generate logic tree samples from the posterior using a learnable GFlowNet, which is a diversity-seeking generator for structured discrete variables. The M-step employs the generated logic rules to approximate marginalization over the posterior, facilitating the learning of model parameters and refining the tunable LLM prior parameters. In the online setting, our locally built, lightweight model will iteratively extract the most relevant rules from LLMs for each sequence using only a few iterations. Empirical demonstrations showcase the promising performance and adaptability of our framework. 
\end{abstract}

\section{Introduction}
\setlength{\abovedisplayskip}{0pt}
\setlength{\abovedisplayshortskip}{0pt}
\setlength{\belowdisplayskip}{0pt}
\setlength{\belowdisplayshortskip}{0pt}
\setlength{\jot}{0pt}
\setlength{\floatsep}{0ex}
\setlength{\textfloatsep}{0ex}
\setlength{\intextsep}{0ex}
\setlength{\topsep}{0ex}
\setlength{\partopsep}{0ex}
\setlength{\parskip}{0.68ex}
Modern systems, such as healthcare, finance, and social media, produce voluminous data that are represented as discrete events with irregular timestamps. Generating concise and human-readable knowledge to explain this intricate event data is of great scientific and practical value. The distilled knowledge can be generalized to other contexts~\citep{ullman2012theory,campero2018logical}. 

For example, in healthcare, electronic health records (EHRs) are often represented as discrete event sequences, containing fine-grained time and type information on doctors' treatments, patients' measurements, and symptoms. It is desirable to generate concise medical knowledge such as the disease phenotypes and therapies, to shed light on these messy events. This will facilitate a deeper understanding of each patient's unique health journey and medical decisions, ultimately leading to more effective and individualized care. 

However, the heterogeneity observed in each patient's data poses a challenge -- each event sequence may exhibit diverse natures of medical histories, treatments, and conditions~\citep{henrich2003evolution,laland2004social}. Generating the most relevant and accurate knowledge to explain such heterogeneous data requires sophisticated methods.
\begin{figure}
% \vskip 0.2in
    \centering
    \includegraphics[width=0.5\textwidth]{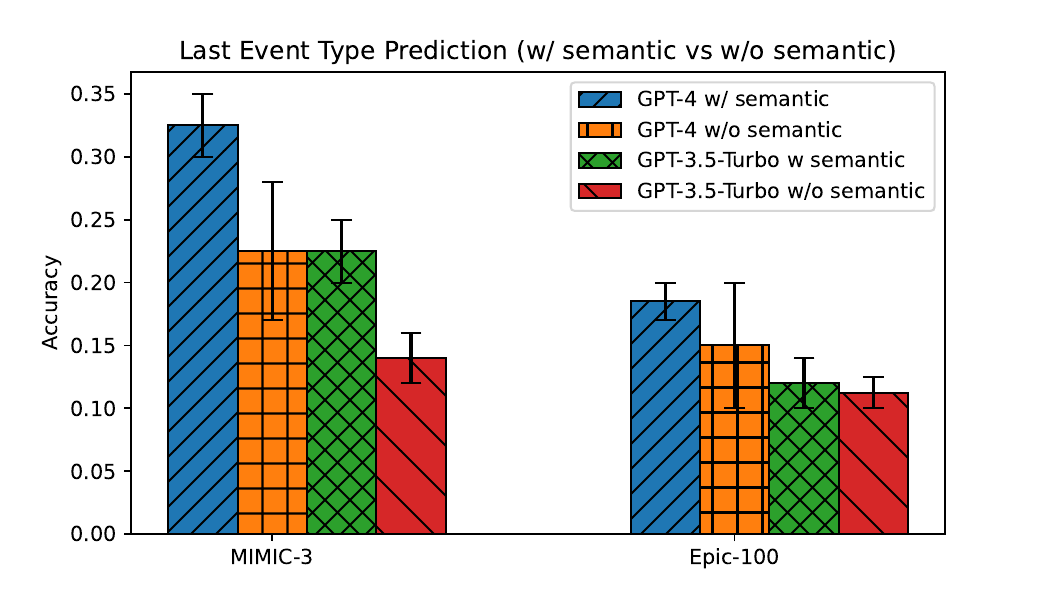}
    \vskip -0.05in
    \caption{\textit{GPT can help last event type prediction}. We found replacing the semantic meaningful event names by numerical event ids in event history degrades the performance of event prediction.}
    \label{fig:gpt_baseline}
\vskip 0.1in
\end{figure}

Recently, Large Language Models (LLMs) have demonstrated promising human-like reasoning abilities as few-shot learners \citep{brown2020language}. When prompted with step-wise explanations of reasoning, these models excel in logical reasoning \citep{pan2023logic}, abstract pattern induction \citep{webb2023emergent}, and social learning \cite{leng2023llm}. Despite their success in text-based reasoning tasks, LLMs still face challenges in extending their reasoning capabilities to tabular data~\citep{hegselmann2023tabllm} and discrete event sequences~\citep{shi2023language}.

In this paper, we propose to leverage LLMs, trained on general-domain data, as a {\it prior} to generate human-readable knowledge. Specifically, we will encourage LLMs to generate logic trees, from their prior distribution $p(\mathcal{R})$. Given the observed discrete event data $\mathbf{X}$, the belief on the logic trees will be updated according to Bayes rule, i.e., $p(\mathcal{R}|\mathbf{X}) \propto p(\mathcal{R}) p(\mathbf{X} |\mathcal{R})$~\citep{leng2023llm,acemoglu2011bayesian}, where we will use a {\it temporal logic point process} (TL-PP)~\cite{li2020temporal} to model $p(\mathbf{X} |\mathcal{R})$. The inference procedure can be regarded as performing the reweighting of each logic tree from LLMs.  The goal of our paper is \textit{to perform the logic tree inference using LLM prior in a tractable and efficient manner}. 

Performing inference on logic trees is challenging since the posterior distribution is intractable due to their discrete combinatorial space. Traditional solutions, such as MCMC, approximate intractable posteriors by sampling, yet is struggling with multi-modal distributions \citep{miao2019cgmh,zhang2020language,lew2023sequential}. Reinforcement learning (RL) methods like proximal policy optimization (PPO) \citep{schulman2017proximal}, treating the sampling process as a policy, may also fail to capture the distribution's full diversity \citep{zhu2023fine}. The problem becomes worse when the target distribution is incorrectly specified, leading to an overoptimized policy. In our paper, we will address the inference challenge using the GFlowNet~\citep{bengio2023gflownet}, a recently proposed sound diversity-seeking generative model for structured discrete variables. As a deep RL algorithm adept at managing unnormalized rewards, GFlowNet has shown effectiveness in fine-tuning Large Language Models with intractable thought posteriors \citep{hu2023amortizing}. We wish to extend its success to Tree-of-Thoughts (ToT) reasoning \citep{yao2023tree}, such as generating logic trees to explain event sequences. 

Our overall learning follows an amortized EM algorithm, where we treat the logic tree as latent variables. In the E-step, we train a GFlowNet to generate logic tree samples from their posterior distribution. The GFlowNet model parameters are shared across all training event sequences, which is the reason why we term it amortized EM. In the M-step, we use the generated logic tree samples to approximate marginalization over the posterior. This process provides an objective function for learning the TL-PP model parameters and refining some tunable LLM parameters (assuming the LLM priors are also learnable). The algorithm iterates between the E-step and M-step until convergence. During the testing stage, given a new event sequence, we employ the trained GFlowNet to efficiently perform inference for explanatory logic trees by sampling from the posterior $p(\mathcal{R}|\mathbf{X}) \propto p(\mathcal{R}) p(\mathbf{X} |\mathcal{R})$, leveraging well-trained priors and models from the training stage. This enables our method to efficiently and adaptively explain previously unseen event sequences.
Our main contributions are: \\ 
({\it i}) We introduce, \textbf{LaTee}, an amortized EM learning framework that can learn to infer and generate \underline{\textbf{La}}tent logic \underline{\textbf{T}}rees to \underline{\textbf{e}}xplain observed \underline{\textbf{e}}vent sequences, which leverages LLMs as prior;  \\
({\it ii}) In the E-step, we use GFlowNets to fine-tune LLMs and enable diverse logic tree generation, which better tackles the heterogeneity issue exhibited for event sequences;\\
({\it iii}) Our method generates a relative margin of 20\% over SOTA Attentioin-based temporal point process (TPP) models on future event prediction based on real-world behavior datasets. This shows that our interpretable and knowledge-driven TPP model is also flexible.

\section{Related Works and Background}
\subsection{Knowledge Extraction from Event Sequences}
Knowledge extraction refers to the process of refining, condensing, or summarizing large volumes of raw data to distill the most relevant and essential information. For noisy event sequences, we will represent our knowledge as a collection of symbolic logic trees, which is a hierarchical and structured representation of logical relationships among different elements or propositions~\citep{campero2018logical}. Our logic tree extraction from events is related to {\it symbolic rule induction} and {\it semantic cognition}. \\
{\bf Symbolic Rule Induction.} Symbolic rule induction refers to the process of automatically discovering logical rules from observed data. Classic symbolic inductive logic programming (ILP) methods \citep{quinlan1990learning,cropper2020logical} mostly adopt discrete search in the space of logic programs and do very well at generalizing from just a few examples. Neuro-symbolic rule inductions \citep{evans2018learning,yang2017differentiable,rocktaschel2017end,campero2018logical} are {\it differentiable} ILP methods and are generally more robust to noisy input. In our approach, we take inspiration from a differentiable \textit{backward chaining algorithm} \citep{rocktaschel2017end} and represent a symbolic logic tree starting from the target predicates. For instance, consider a \textit{Put-into} task as our target predicate, in which we need to replace element $X$ from box $Y_{1}$, room $Z_{1}$ into box $Y_{2}$, room $Z_{2}$. We can represent this actionable strategy as a set of ordering logic rules as:
 \begin{align}
 \textit{Put-into}(X, Y) &\leftarrow \textit{Open}(Y) \wedge \textit{Pick-up}(X), \\
  \textit{Pick-up}(X) &\leftarrow  \textit{Open}(Y) \\
   \textit{Open}(Y) &\leftarrow \textit{Move-to}(Z)
 \end{align}
which is a logic tree, with $ \textit{Put-into}(X, Y)$ being the root and other predicates being its children. Many classic or differentiable ILP methods can automatically learn such rules from data, however, they require carefully hand-crafted rule templates for each ILP task in order to constrain and reduce the search space effectively~\citep{glanois2022neuro}.\\
{\bf Semantic Cognition.} Semantic cognition refers to the development of systems that can comprehend and manipulate meaning in a manner similar to human cognitive processes. It explores how knowledge is organized, represented, and utilized to derive semantic understanding from various forms of data. Previous research has described it as a process similar to reducing logical dimensions \citep{katz2008modeling,ullman2012theory} through employing probabilistic generative models. These models are capable of learning both logical rules and fundamental relationships that explain the data observed. Similar to ILP methods, they can perform deductive reasoning using logical rules. However, unlike traditional ILP methods, these models can also induce facts. While these approaches showed potential, they faced significant issues with scalability. The recent advancements in social learning in LLMs \citep{leng2023llm} suggest that it might be beneficial to reexamine these concepts.
\subsection{Knowledge-Driven Probabilistic Models for Event Sequences}
Temporal point process (TPP) provides an elegant probabilistic model for continuous-time event sequences, which is characterized by an intensity function.  The intensity function represents the occurrence rate of events, which is usually modeled as parametric, nonparametric, or deep neural network forms. Traditional parametric TPP models like the Hawkes process offer interpretability, but their simplicity limits flexibility. On the other hand, neural-based models, such as RMTPP~\cite{du2016recurrent} and Transformer Hawkes~\cite{zuo2020transformer}, provide expressiveness but are often criticized for their black-box nature and hinder their applications in high-stakes scenarios. In this paper, we aim to generate logic trees from a fine-tuned LLM to inform the functional form of the intensity, which strikes a balance between model flexibility and interpretability. The modeling idea takes inspiration from TL-PP~\cite{li2020temporal} , as detailed below. \\
{\bf Rule-informed Event Sequences.} We will build a rule-informed conditional intensity function for the event sequences as:
\begin{align}
\label{eq:intensity}
    \lambda(t; w, \mathcal{R}, \mathbf{X}_{t}) := \text{exp}\Big\{\sum_{f \in \mathcal{R}} w_{f} \phi_{f}(\mathbf{X}_{t}) + b(t)\Big\}, 
\end{align}
where $f$ is a valid path from the symbolic logic tree $\mathcal{R}$, $\phi_{f}(\mathbf{X}_{t})$ is the logic-informed feature derived from the number of ordered event combinations in event history $\mathbf{X}_{t}$ satisfying the path $f$ (with more details can be found in ~\cite{li2020temporal}), and $w_{f}$ is the weight corresponding to rule $f$. Given this probabilistic model, we can use the negative log-likelihood of the temporal point process as a loss function to jointly learn weights $w$ and symbolic structure $\mathcal{R}$. Given a event sequence $\mathbf{X}=\{(t_{i}, e_{i})\}_{i=1}^{L}$ observed over an interval $[0, T]$, the negative log-likelihood of $\mathbf{X}$ is expressed as:
\begin{align}
\label{eq:llh}
    &\mathcal{L}_{w, \mathcal{R}}(\mathbf{X}) = \\ & -\sum_{j=1}^{L}\log \lambda(t_{j} ; w, \mathcal{R}, \mathbf{X}_{j}) + \int_{0}^{T} \lambda(t ; w, \mathcal{R}, \mathbf{X}_{t}) dt. \notag
\end{align}
where each $t_j$ is the event trigger time and $\mathbf{X}_{j}$ refers to the event sequences up to $t_j$. Nevertheless, this learning problem is challenging because it requires learning the parameters $w$ in a continuous space as well as the symbolic structure $\mathcal{R}$ in a discrete space.

\begin{figure*}
    \centering
    \includegraphics[width=\textwidth]{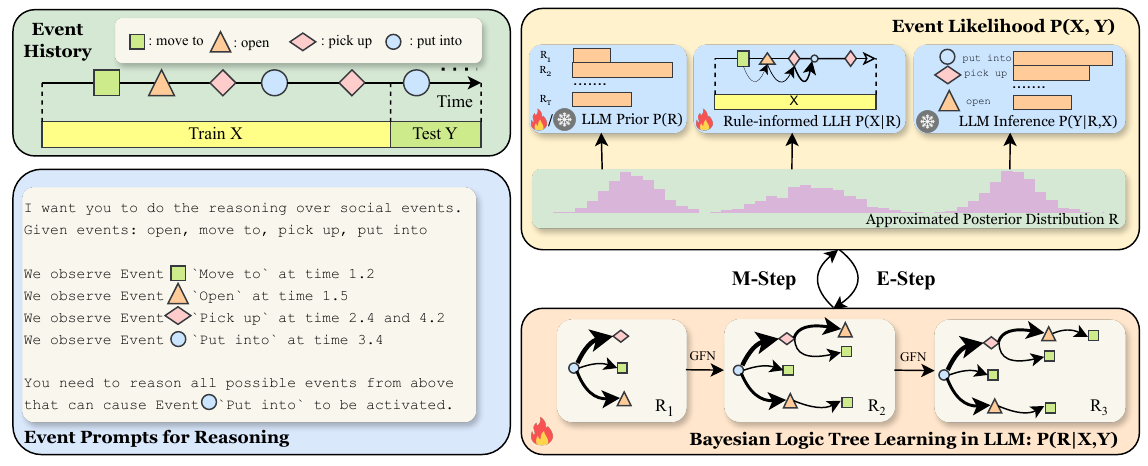}
    \vskip -0.1in
    \caption{\textit{The Architecture of Proposed Framework}. The presented event history represents a typical human trajectory: beginning with relocating to a new place, opening a box, picking up an object, and culminating in placing the retrieved item in a specific location. In the training phase, we first convert this explicit event history into a textual format. Subsequently, we employ a LLM to execute conditional sampling and perform backward reasoning, starting from the predetermined goal (E-Step). The resultant reasoning pathway is then transformed into a symbolic logic tree, which aids in updating the event probabilities (M-Step). In this context, the \textit{campfire} icon signifies that the model is being updated, while the \textit{snowflake} icon indicates that the model is in a `frozen' state. We use the thickness of a path in the symbolic logic tree to represent its posterior probability.}
    \label{fig:arch}
\vskip -0.2in
\end{figure*}

% \textbf{Semantic Cognition.}
% Semantic cognition deals with how knowledge is acquired and integrated. Previous research has described it as a process similar to reducing logical dimensions \citep{katz2008modeling,ullman2012theory} through employing probabilistic generative models. These models are capable of learning both logical rules and fundamental relationships that explain the data observed. Similar to ILP methods, they can perform deductive reasoning using logical rules. However, unlike traditional ILP methods, these Bayesian models can also induce facts. While these approaches showed potential, they faced significant issues with scalability. The recent advancements in social Bayesian learning in Large Language Models \citep{leng2023llm} suggest that it might be beneficial to reexamine these concepts.

\subsection{Human-Like Reasoning in LLMs}
Recent developments in LLMs, such as GPT-4 \citep{achiam2023gpt} and LlaMA 2 \citep{touvron2023llama}, have extended the capabilities of AI beyond conventional predictive analytics to simulate sophisticated human-like interactions in various systems~\citep{gao2023s}. In-context learning (ICL) within LLMs is a notable feature where the model performs tasks based on input-output examples without adjusting any parameters. Importantly, ICL can be understood through a Bayesian inference framework \citep{xie2021explanation}, where the augmented prompt serves as a semantic prior, guiding \textit{latent} concepts acquired during pre-training for chain of thought reasoning and subsequent output \citep{wei2022chain,kojima2022large}. Despite the transformative nature of ICL in LLMs, which allows them to adapt to new tasks without explicit retraining, challenges remain in explaining extrapolation to unseen tasks and understanding the impact of model architecture and optimization. Conversely, knowledge extraction from locally deployable LLMs, achieved through careful fine-tuning \citep{schick2020s} using Parameters Efficient Fine Tuning (PEFT) techniques \citep{hu2021lora,dettmers2023qlora}, also provides valuable insights. We will adopt PEFT ideas in this paper. 
\subsection{Intractable Bayesian Inference in LLMs}
% The intractable inference problem emerges from LLM since next-token conditional prediction in autoregressive language models decomposes the distribution over sequences of tokens as a product of $p(w_{1:N}) = p(w_{1})p(w_{2}|w_{1}) \cdots p(w_{N}|w_{1:N-1})$. While this decomposition makes left-to-right sampling from the distribution tractable, sampling from other conditional distributions is intractable. For instance, 
The challenge in inferring the latent logical reasoning path from LLMs stems from the intractability of the posterior. Given question-answer pair $(X, Y)$, the posterior of the latent chain-of-thought $p_{LM}(Z |X, Y) = \frac{p_{LM}(X, Z, Y)}{\sum_{Z'}p_{LM}(X, Z', Y)}$ is intractable due to the discrete combinatorial space for thoughts $Z$ ~\citep{hu2023amortizing}. Existing approaches to address this intractable inference problem in language models often resort to tokenwise approximations using techniques like tempered and contrastive sampling \citep{malkin2021coherence,li2022contrastive}, along with problem-specific strategies like beam search and local search techniques \citep{lu2021neurologic,sha2020gradient}. In our paper, we will use GFlowNets to guide posterior sampling of logic trees and fine-tune LLMs.
% The intractable inference problem emerges from LLM since next-token conditional prediction in autoregressive language models decomposes the distribution over sequences of tokens as a product of $p(w_{1:N}) = p(w_{1})p(w_{2}|w_{1}) \cdots p(w_{N}|w_{1:N-1})$. While this decomposition makes left-to-right sampling from the distribution tractable, sampling from other conditional distributions is intractable. For instance, given a question-answer pair $(X, Y)$, the posterior of latent chain-of-thought $p_{LM}(Z |X, Y) = \frac{p_{LM}(Z, X, Y)}{\sum_{Z'}p_{LM}(Z',X, Y)}$ is intractable \citep{hu2023amortizing}. Current approaches to the intractable inference problem in language model use tokenwise approximations by tempered and contrastive sampling \citep{malkin2021coherence,li2022contrastive} and problem-specific beam search and local search techniques \citep{lu2021neurologic,sha2020gradient}.

\textbf{GFlowNets as Posterior Samplers in LLMs.}
GFlowNets \citep{bengio2021flow,bengio2023gflownet} are originally introduced as a diversity-seeking probabilistic reinforcement learning algorithm for molecular discovery. A recent work \citep{hu2023amortizing} connects GFlowNets to Chain-of-Thoughts (CoT) generation, by leveraging the amortized inference ability of GFlowNets. In their work, given an unnormalized density (reward) $R:\mathcal{Z} \to \mathbb{R}_{>0}$, GFlowNets learn policies to sample sequences in a token-level, i.e., $\mathbf{Z}_{t} = z_{1} z_{2} \cdots z_{t} \texttt{T} \in \mathcal{Z}$ (where $z_{i}$ is a language token and $\texttt{T}$ denotes End of Sentence token), as if they were sampling from a target distribution. The goal of GFlowNets is to fine-tune a token-level language generation  $q_{GFN}(\mathbf{Z}_{t} | \mathbf{Z}_{t-1}; \theta)$ initialized by LLM such that marginal $q_{GFN}(\mathbf{Z}_{t}) \propto r(\mathbf{Z}_{t})$, i.e., driving the marginal likelihood of generating a complete sequence is proportional to its reward. The learning objective for GFlowNets is defined by the subtrajectory balance (SubTB) objective, equivalent to the path consistency objective \citep{nachum2017bridging, deleu2024discrete,tiapkin2024generative} in Max-Entropy RL \citep{haarnoja2017reinforcement}. 

% This objective has also been previously employed in the context of text generation \citep{guo2021efficient}.
\section{Our Proposed LaTee using LLM prior}
Instead of eliciting linear thoughts from LLMs, our focus is to extract and reweight symbolic logic trees generated from LLMs to explain the dynamics of the observed event sequences. We hope the obtained logic trees will not only offer personalized explanations for each event sequence but also enable accurate future events prediction.

\subsection{LLM-Symbolic Integration by Latent Variables}
Given event sequences $X$ and next event type $Y$, where $X$ records explicit event sequences with $k$ events $X = \{(t_{i}, e_{i})\}_{i=1}^{k}$, where $t_i$ is the $i$-th event time, $e_{i}$ is the $i$-th event type, and $Y= e_{k+1}$ is the next event type after the $k$-th event. We are interested in finding a collection of latent symbolic logic trees $\mathcal{R}$, which are composed of various event types that trigger subsequent events and best explain the likelihood of the observed event sequence:
\begin{equation}
    p(X, Y) = \sum_{\mathcal{R}} p(X, Y | \mathcal{R} ) p(\mathcal{R}).
\label{eq:mix_model}
\end{equation}
For this mixture latent variable model (LVM), we treat $\mathcal{R}$ as latent variables; $p(\mathcal{R})$ is the prior distribution for the latent logic tress; and the joint likelihood of the event sequences $p(X, Y | \mathcal{R} )$ can be derived from temporal point process framework, as shown in Eq.~(\ref{eq:llh}).

We will employ LLMs as the prior $p(\mathcal{R})$ for logic trees. Additionally, if we aim to leverage the powerful reasoning and generation capabilities of LLMs to predict $Y$ --- for instance, in the context of symptom-treatment pairs or question-answer pairs for $(X,Y)$ --- it becomes intriguing to explore the recommendations of $Y$ given $X$ provided by LLMs. Consequently, we further decompose the mixture LVM (as shown in Eq.~(\ref{eq:mix_model})) as:
\begin{align}
\label{eq:joint_llh}
    p(X, Y) &= \sum_{\mathcal{R}}p(X, Y | \mathcal{R}) p_{LM} (\mathcal{R}), \\ 
    &= p_{LM}(Y | X, \mathcal{R}) \sum_{\mathcal{R}}p_{w}(X | \mathcal{R}) p_{LM} (\mathcal{R}; \phi).
\end{align}
We aim to jointly optimize the event likelihood parameter $w$ and the tunable parameters $\phi$ in the prior language model. The challenge in learning arises from the latent variables $\mathcal{R}$. Fortunately, the EM algorithm provides an effective tool for learning mixture models with latent variables. However, in the E-step, we need to analytically evaluate the current posterior $p(\mathcal{R} | X, Y) \propto p_{LM}(Y | X, \mathcal{R}) p_{w}(X | \mathcal{R}) p_{LM} (\mathcal{R})$, which is intractable due to that the partition function requires the summation over the discrete space of $\mathcal{R}$. To tackle this intractability, variational-EM algorithm \citep{dempster1977maximum, beal2003variational, koller2009probabilistic} can be used to approximate the posterior by optimization. We will address this issue by introducing an amortized EM, where in the E-step we learn GFlowNets to sample from $p(\mathcal{R} | X, Y) $ without the need to calculate the partition function. 
\subsection{Amortized EM framework for Logic Tree Inference}
The derivation of the Evidence Lower Bound (ELBO) for Eq.~(\ref{eq:joint_llh}) is presented in Appendix \ref{appsec:elbo}. It explains the rationale for analytically evaluating the posterior in the E-step to achieve a tight ELBO. 

Specifically, in the E-step, we will draw samples from the posterior over the latent symbolic logic tree, denoted as $p_{LM}(\mathcal{R} | X, Y)$, which comes from an amortized sampler of $\mathcal{R}$ with an LLM as its policy.
In the M-step, we maximize the log-likelihood of the joint probability of the sampled latent variables $\mathbb{E}_{\mathcal{R} \sim p(\mathcal{R}|X, Y)}[\log  p_{LM}(Y | X, \mathcal{R}) p_{w}(X |  \mathcal{R}) p_{LM}(\mathcal{R}) ]$ with respect to the parameters of $w$. This combination of amortized inference (learning to sample the symbolic logic tree from the language model) and supervised learning (optimizing the likelihood model with the ‘supervision’ involving $\mathcal{R}$ sampled from the amortized posterior) is presented in Fig. \ref{fig:arch}. We illustrate them in detail in the sections below.

\textbf{E-Step: Amortized Inference with GFlowNets.}
For inference in the high-dimension discrete latent space, we leverage the probabilistic framework of GFlowNets \citep{bengio2021flow,bengio2023gflownet}. Consider a symbolic logic tree $\mathcal{R}$, we start from the root $\mathcal{R}_{0} := \{z_{0}\}$, in which $z_{0}$ is the target predicate (can be composed by multiple language tokens). We follow \textit{backward chaining} \citep{rocktaschel2017end} to form a symbolic proof tree in a top-down fashion by prompting LLMs. We grow the logic tree one level deeper at a time based on the previous paths. Concretely, suppose $\mathcal{R}_{t}$ can be represented by $m$ paths, \textit{i.e.,} $\mathcal{R}_{t} := \{z_{0}^{(i)}z_{1}^{(i)}\cdots z_{j}^{(i)}\}_{i=1}^{m}$, where $z_{j}^{(i)} \in \mathcal{Z}$ is the $j$-th predicate presented in the $i$-th path from the predefined predicate space $\mathcal{Z}$. If the maximum number of nodes for each path to expand is constrained to $W$, the generative process from a symbolic logic tree $\mathcal{R}_{t}$ to $\mathcal{R}_{t+1}$ can be represented as:
\begin{equation}
  \log  q_{GFN} (\mathcal{R}_{t+1} | \mathcal{R}_{t}) := \sum_{i=1}^{m}\sum_{k=1}^{W+1}  \log q_{LM} (z_{j+1}^{(i),k} | z_{0}^{(i)} \cdots z_{j}^{(i)}),
\end{equation}
where $q_{LM}$ is the autoregressive sequence generation model and $z_{j+1}^{(i),k}$ is the next level predicates chosen from $\mathcal{Z}_{W} \cup \{\texttt{T}\}$, $\mathcal{Z}_{W} \subseteq \mathcal{Z}$, $|\mathcal{Z}_{W}|=W$, and \texttt{T} denotes a stop symbol. The nodes in the symbolic logic tree thus grow in $O(W^{n})$ and will not stop expanding until all the paths reach the termination state $\mathtt{T}$, i.e., 
\begin{equation}
    \log q_{GFN} (\texttt{T} | \mathcal{R}_{t}) :=  \sum_{i=1}^{m}\log q_{LM} (\texttt{T} | z_{0}^{(i)} \cdots z_{j}^{(i)}).
\end{equation}
The marginal likelihood of sampling a terminal logic tree $\mathcal{R}_{t}$ is given by 
\begin{align*}
    &q_{GFN}(\mathcal{R}_{t} \to \mathtt{T})  = \\  &\int_{\tau=(\mathcal{R}_{0} \leadsto \mathcal{R}_{t})} \Pi_{i=1}^{t} q_{GFN}(\mathcal{R}_{i} | \mathcal{R}_{i-1}) q_{GFN}(\texttt{T}|\mathcal{R}_{t}) d \tau
\end{align*}
% $$q_{GFN}(\mathcal{R}_{t} \to \mathtt{T})  = \\  \int_{\tau=(\mathcal{R}_{0} \leadsto \mathcal{R}_{t})} \Pi_{i=1}^{t} q_{GFN}(\mathcal{R}_{i} | \mathcal{R}_{i-1}) q_{GFN}(\texttt{T}|\mathcal{R}_{t}) d \tau$$
over trajectories $\tau$ starting at $\mathcal{R}_{0}$ and ends at $\mathcal{R}_{t}$.  Notably, the goal of GFlowNet training is to fit the parametric policy $q_{GFN}(\cdot | \cdot; \theta)$ such that its terminating probability $q_{GFN}(\mathcal{R}_{t} \to \mathtt{T})$ is proportional to a predefined reward $r$. In our case, GFlowNet's reward $r$ is defined as the posterior of logic trees $p(\mathcal{R}|X, Y)$, i.e., $r(\mathcal{R} | X, Y) \propto p_{LM}(Y | X, \mathcal{R}) p_{w}(X | \mathcal{R}) p_{LM} (\mathcal{R})$. By construction, GFlowNet's marginal terminating distribution is proportional to its reward function $r(\mathcal{R} | X, Y)$, thus we will have the final samples $\mathcal{R}$ given by the GFlowNet's policy $q_{GFN}(\cdot | \cdot, \theta)$ following the distribution of unnormalized true posterior of $p(\mathcal{R} | X, Y)$. Here, the given reward function $r$ can be decomposed as a product of likelihood terms that accumulated over steps of the sampling sequence. In this case, a forward-looking SubTB loss \citep{madan2023learning} for GFlowNet can help local credit assignment \citep{hu2023amortizing,hu2023gflownet}. The SubTB learning objective for trajectory $\tau = (\mathcal{R}_{0}, \mathcal{R}_{1}, \cdots, \mathcal{R}_{t})$ is:
% {
% \small\begin{equation}
% \label{eq:subtb}
%     \mathcal{L}(\mathcal{R}_{t})=\sum_{0 \le i < j \le t} \Bigg( \log \frac{R(\mathcal{R}_{i}^{\texttt{T}})\Pi_{k=i+1}^{j}q_{\theta}(\mathcal{R}_{k}|\mathcal{R}_{k-1})q_{\theta}(\texttt{T}|\mathcal{R}_{j})}{R(\mathcal{R}_{j}^{\texttt{T}})q_{\theta}(\texttt{T}|\mathcal{R}_{i})}\Bigg)^{2},
% \end{equation}
% }
\begin{align}
\label{eq:subtb}
&\mathcal{L}_{SubTB}(\theta)=\\&\sum_{0 \le i < j \le t}  \left[ \log \frac{r(\mathcal{R}_{i}^{\texttt{T}})\Pi_{k=i+1}^{j}q_{\theta}(\mathcal{R}_{k}|\mathcal{R}_{k-1})q_{\theta}(\texttt{T}|\mathcal{R}_{j})}{r(\mathcal{R}_{j}^{\texttt{T}})q_{\theta}(\texttt{T}|\mathcal{R}_{i})}\right]^{2},\notag
\end{align}
where $q_{\theta}$ is the conditional GFlowNet policy initialized by a language model $p_{LM}$ conditioned on prefix $X$ and $Y$. The detailed derivation of the SubTB loss is given in Appendix \ref{appsec:subtb}. In practice, this loss can be minimized by gradient descent on $\theta$ sampled either \textit{on-policy} or \textit{off-policy}, just as in reinforcement learning. To predict the event type $Y$ for an unseen event sequence $X$, one can draw samples of $\mathcal{R}$ from $q_{\theta}$ followed by sampling from $p_{LM}(Y | X, \mathcal{R})$.
\begin{algorithm}[tb]
   \caption{Bayesian Logic Tree Learning for Events}
   \label{alg:latee}
\begin{algorithmic}
   \STATE {\bfseries Input:} data pool  $\{\mathcal{X}, \mathcal{Y}\}$, rule weights $w$, tunable parameters $\theta$ for LLM as the GFlowNet policy, tunable parameters $\phi$ for LLM as the prior policy, optimization and exploration hyperparameters, threshold $\alpha$
   \REPEAT
   \STATE sample batch data pair $(X, Y) \sim \{\mathcal{X}, \mathcal{Y}\}$
   \STATE sample $\tau \sim q_{\theta}(\tau | X, Y)$; $\tau = (\mathcal{R}_{0},\cdots, \mathcal{R}_{T})$
   \STATE $r_{t} \leftarrow  p_{w}(X | \mathcal{R}_{t})p(Y | X, \mathcal{R}_{t})p_{\phi} (\mathcal{R}_{t}), t= 0, \cdots, T$
   \STATE $\mathcal{L}_{SubTB} \leftarrow$ SubTB loss in Eq. (\ref{eq:subtb}) along $\tau$ with reward $r_{t}$
   \STATE E-step: GD on $\theta$ with $\nabla_{\theta} \mathcal{L}_{SubTB}$
   \IF{$\mathcal{L} < \alpha $}
   \STATE Sample $\tau \sim q_{\theta}(\tau | X, Y)$
   \STATE M-step: GD on $w$ and $\phi$ with $\nabla_{w,\phi} \mathcal{L}_{llh} $ in Eq. (\ref{eq:likelihood}) 
   \ENDIF
   \UNTIL{some convergence criteria}
\end{algorithmic}
\end{algorithm}

\begin{figure*}[ht]
    \centering
\includegraphics[width=\textwidth]{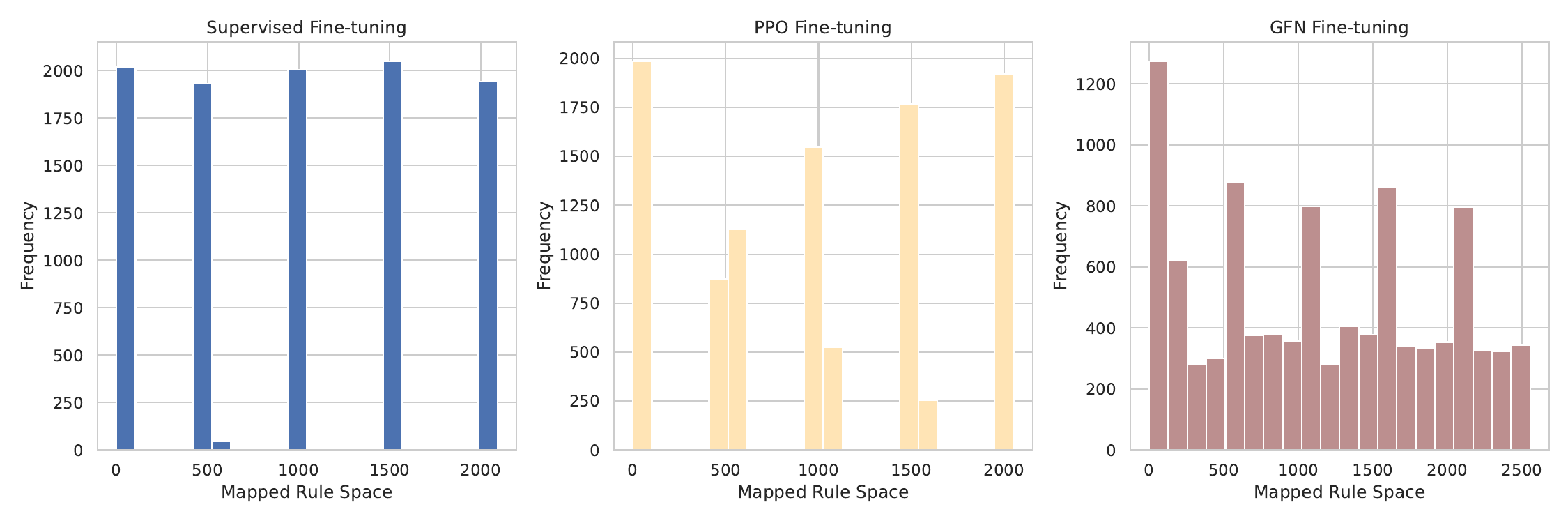}
    \vskip -0.1in
    \caption{Empirical rule distributions sampled from the Language Model fine-tuned by three different approaches. We use 10,000 samples to depict the frequency distribution of a complete logic rules search space with support $|\mathcal{R}|=2500$. The x-axis represents these logic rules in a nominal 1-D format, where each point corresponds to a specific rule. The ordering of these points is not indicative of any inherent sequence.}
    \label{fig:diversity}
\vskip -0.1in
\end{figure*}

\textbf{M-Step: Model Parameter Updating.}
The marginal terminal distribution of GFlowNet is used as a variational approximation to the intractable posterior $p(\mathcal{R}|X,Y)$ to perform updates to the generative model's parameters. Thus, for the event demonstrations $X$ and next event type $Y$, we can uncover its underlying symbolic logic tree $\mathcal{R}$ from the policy of the conditional GFlowNet $q_{GFN}$ and perform in expectation gradient update on the parameters $w$ of event likelihood and tunable parameters $\phi$ for structure prior learning:
% \begin{align}
% \label{eq:likelihood}
% \begin{split}
%     \mathcal{L}(w, \phi) = \mathbb{E}_{\mathcal{R} \sim q(\mathcal{R} | X,Y)}[\log p_{w}(X | \mathcal{R}) + \\ \log p_{LM}(Y | X, \mathcal{R}) + \log p_{\phi}(\mathcal{R})].
% \end{split}
% \end{align}
\begin{align}
\label{eq:likelihood}
    \notag
    \mathcal{L}_{llh}(w, \phi) =  \mathbb{E}&_{ \mathcal{R} \sim q_{GFN}(\mathcal{R} \to \mathtt{T})}[\log p_{w}(X | \mathcal{R}) \\ &+\log p_{LM}(Y | X, \mathcal{R}) + \log p_{\phi}(\mathcal{R})].
\end{align}
It should be noted that the evolving nature of the generative models $p_{w}$, $p_{LM}$, and $p_{\phi}$ during joint optimization leads to a dynamic reward system for the GFlowNets. The training process involves alternating between E-steps and M-steps, with the frequency of GFlowNet updates between successive M-steps being a variable parameter that can be either predetermined or adaptively chosen. Following the approach outlined in \citep{hu2023gflownet}, adaptive E-steps are implemented through loss thresholding. This method uses a moving average of the GFlowNet’s training loss as a measure of the accuracy in approximating the true posterior. An M-step gradient update is executed following a GFlowNet update only if this moving average drops below a set loss threshold. The overall algorithm is presented in Alg. \ref{alg:latee}.

\subsection{Discussion}
{\bf Comparison with ILP systems.} In our approach, we harness LLMs to generate latent logic trees, replacing traditional symbolic ILP systems. While traditional ILP systems that are based on discrete space search excel in rule learning from minimal examples, they are sensitive to noisy input, and a single error can lead to malfunction. Neural-symbolic rule induction systems, on the other hand, are more robust to noise but will struggle with few-shot learning and may face the risk of overfitting. Symbolic reasoning through LLMs integrates pretrained knowledge, addressing the challenge of learning rules from limited data. Additionally, our approach employs a step-by-step process reflecting human cognitive functions~\citep{wei2022chain}, enhancing semantic understanding from event sequences.\\
{\bf Comparison with GFlowNets-CoT.} We share some similarities with GFlowNets-CoT \citep{hu2023amortizing} in the Bayesian inference stage in which LLM is used as a probabilistic generative model that simultaneously generates logical rules and a set of core relations underlying them. However, the distinction between our approach and GFlowNets-CoT is that we extend GFlowNets fine-tuning to a sentence-level symbolic logic tree $\mathcal{R}$, which is similar to sentence-level Tree-of-Thoughts \citep{yao2023tree},  through directly querying sentence probabilities within a confined `sentence space' for backward learning, as opposed to ToT's forward-only inference approach.
GFlowNets-CoT's likelihood model relies only on $p_{LM}(X, Y, \mathcal{R})$ while ours is built upon a \textit{modular} likelihood model by decomposing $p(X, Y | \mathcal{R})$ into event likelihood $L_{w}$ along with the language likelihood $p_{LM}$, $\textit{i.e.,}$ $p_{LM}(\mathcal{R})p_{w}(X | \mathcal{R})p_{LM}(Y | X, \mathcal{R})$.

\section{Experiments}

\begin{table*}[ht]
\caption{Last event prediction performance on three real-world behavior datasets using both attention-based TPP models and Language Model \textit{Opt-1.5B} as predictors. ER stands for Error Rate and MR stands for Mean Rank. The performance is averaged over three different seeds and the standard deviation is stored in the parenthesis. The best performance is in bold and also highlighted in gray.}
\label{tab:main_res}
\vskip 0.15in
\begin{center}
\scalebox{0.85}{
\begin{sc}
\begin{tabular}{lcccccccr}
\toprule
  & Dataset & \multicolumn{2}{c}{MIMIC-3} & \multicolumn{2}{c}{Epic-100}  & \multicolumn{2}{c}{StackOverflow} \\ \cmidrule{3-8} 
Method & Metrics & ER (\%) $\downarrow$   & MR $\downarrow$  &ER (\%) $\downarrow$ & MR $\downarrow$ & ER (\%) $\downarrow$ 
 & MR $\downarrow$\\
 \midrule
\multicolumn{2}{l}{AttNHP \citep{yang2021transformer}} 
& 36.66(13.76) & 1.25(0.00) & 67.53(0.00) & 2.45(0.00) & \cellcolor{gray!50} \textbf{33.33}(0.00) & \cellcolor{gray!50}\textbf{1.95}(0.10) \\
\multicolumn{2}{l}{Pt-AttNHP \citep{xue2023prompt}} 
& 77.50(0.00) & 1.75(0.00) & 68.33(1.44) & 2.27(0.16) & 71.11(32.71) & 3.40(0.81) \\
\midrule
\multirow{3}{*}{$k$-shot CoT} & $k$ = 0 
& 100.00(0.00) & 2.24(0.02)  & 78.75(1.76) & 4.75(0.00) & 100.00(0.00) & 3.33(0.00) \\ 
& $k$ = 1 
& 100.00(0.00) & 2.23(0.00) & 76.25(1.57) & 4.66(0.02) & 100.00(0.00) & 3.33(0.00)\\
& $k$ = 3 
&  100.00(0.00) & 2.23(0.00) & 76.25(0.02) & 4.63(0.00) & 100.00(0.00) & 3.13(0.00) \\
\multicolumn{2}{l}{ToT (depth$=3$, width$=3$)} 
& 100.00(0.00) & 2.23(0.00) & 71.25(1.76) & 4.69(0.12) & 96.67(4.71) & 3.07(0.09) \\
\multicolumn{2}{l}{SFT fine-tuning} 
& 82.50(2.50) & 2.14(0.03)  & 75.83(1.44) & 4.38(0.11) & 93.33(6.67) & 4.44(0.30) \\
\multicolumn{2}{l}{PPO fine-tuning} 
& 77.50(0.00) & 2.55(0.00) & 77.50(0.00) & 3.99(0.03) & 73.33(6.67) & 3.29(0.32)\\
\multicolumn{2}{l}{GFN fine-tuning} 
& \cellcolor{gray!50}\textbf{27.50}(8.66) & \cellcolor{gray!50}\textbf{1.14}(0.05) & \cellcolor{gray!50}\textbf{55.25}(9.01) & \cellcolor{gray!50}\textbf{2.15}(0.48) & 33.45(5.12) & 2.23(0.23) \\
\bottomrule
\end{tabular}
\end{sc}
}
\end{center}
\vskip -0.1in
\end{table*}

\subsection{Experimental Setup}
\textbf{Datasets and Evaluation Setup.}
Our study involves one synthetic and three real-world event sequence datasets, containing both semantic and non-semantic information. We view events in these datasets as predicates that can form a symbolic logic tree. For each sequence, we focus on predicting the final event. For the synthetic dataset, we create sequences of event predicates sampled from a prespecified TL-PP using the thinning algorithm \citep{ogata1981lewis}. The functional form of the intensity is informed by the predefined logic rules. Regarding the real-world datasets, one is the \textbf{MIMIC-III} \citep{johnson2016mimic}, an electronic health record dataset from intensive care unit patients. We use various lab measurements and treatment approaches as event predicates. The other is \textbf{EPIC-KITCHENS-100} (EPIC-100) \citep{Damen2021PAMI}, which documents everyday kitchen activities from a first-person perspective over several days, with actions labeled. We analyze these labeled actions in sequence to predict the human's next action based on their past activities. The final one is \textbf{StackOverflow} (SO)  \citep{leskovec2014snap}, which records a sequence of reward history with badges from the question-answering website \textit{StackOverflow} to promote the engagement among its users. Each event in the sequence signifies the receipt of a particular metal. For all the datasets, We consider each sequence as a record pertaining to a single individual and partition each dataset into 80\%, 10\%, 10\% train/dev/test splits by the total population. More details about these datasets can be found in the Appendix \ref{appsec:dataset}.

\textbf{Metrics.}
We follow the common next-event prediction task in TPPs \citep{du2016recurrent, mei2017neural} and emphasize the performance of last event type prediction $k$ from its history $\mathcal{H}$ output by the language model. We evaluate the prediction $\hat{k}$ by \textit{Error Rate} (ER) and \textit{Mean Rank} (MR) that measures the average rank of the ground-truth type in the list; a smaller MR means a higher rank, and thus a better result.
 
\textbf{Base models.}
In this study, we utilize three distinct sizes of language models from the OPT family \citep{zhang2022opt}: \textit{Opt-125M} (small), \textit{Opt-1.5B} (medium), and \textit{Opt-6.7B} (large), as our foundational language model backbones for latent logic tree extraction $p_{LM}(\mathcal{R}|X)$ in the \textbf{E-steps}. These models are fine-tuned for logic tree learning using the LoRA adaptation layer and further optimized through quantization  \citep{dettmers2023qlora} to minimize GPU memory consumption during both forward and backward processing stages. We use \textit{Zephyr-3B} \citep{tunstall2023zephyr} and \textit{Mistral-7B-Instruct} \citep{jiang2023mistral} as frozen inference models for $p_{LM}(Y|X,\mathcal{R})$ in the \textbf{M-steps}.  Detailed methodologies and specifics regarding the fine-tuning of these Large Language Models (LLMs) can be found in the Appendix \ref{appsec:ft_llm}.

\textbf{Competitors.} In our study, we categorize competitors into three distinct types. The first category includes prompt-based approaches applied to language models, such as $k$-shot Chain-of-Thought (\textbf{CoT}) \citep{wei2022chain} and Tree-of-Thoughts (\textbf{ToT}) \citep{yao2023tree}, which are utilized to generate reasoning chains and make prediction of the last event. The second category involves fine-tuning methods for language models, notably supervised fine-tuning (\textbf{SFT}) and Proximal Policy Optimization (\textbf{PPO}) \citep{schulman2017proximal} fine-tuning. The final category consists of advanced neural Temporal Point Process (TPP) models specifically designed for event prediction. Within this group, we focus on \textbf{AttNHP} \citep{yang2021transformer}, an attention-based TPP whose performance is either on par with or superior to the Neural Hawkes Process (NHP) \citep{mei2017neural} and other attention-based models \citep{xue2023easytpp}. Additionally, we consider PromptTPP \citep{xue2023prompt}, a prompting model based on AttNHP (abbreviated as \textbf{Pt-AttNHP}), tailored for processing streaming events with a retrieval memory mechanism. 

\begin{table}[t]
\vskip -0.1in
\caption{\textit{Scalability of the proposed model on two real-world behavior datasets.} The tree traversal depth $d$ and tree expansion width $w$ is fixed to $3$ without further clarification in the Table. We use \textit{Opt-1.3B} for Epic-100 and Stackoverflow as base model in E-step. Error Rate (\%) is used as the evaluation metric for both Epic-100 and StackOverflow. The performance is averaged over three different seeds and the standard deviation is stored in the parenthesis.}
\label{tab:scal}
\begin{center}
\scalebox{0.8}{
\begin{sc}
\begin{tabular}{lcccr}
\toprule
  & Dataset &  \multirow{2}{*}{\makecell{Epic-100 \\ w/ sc}} & \multirow{2}{*}{\makecell{SO \\ w/o sc}} \\
Method &  &  & \\
\midrule
\multirow{3}{*}{\shortstack{LaTee \\(\textit{Tree Depth})}} 
& $d$ = 2 & 69.23(9.43)&  73.31(23.29)\\ 
& $d$ = 3 & 69.40(9.21) & 34.50(5.23) \\
& $d$ = 4 & 68.21(8.13) & 34.41(5.08)\\
\midrule
\multirow{3}{*}{\shortstack{LaTee \\ (\textit{Tree Width})}} 
& $w$ = 3 & 69.40(9.21)& 34.50(5.23) \\ 
& $w$ = 5 & 61.31(9.24) & 33.53(5.10) \\
& $w$ = 7 & 55.25(9.01)&  34.37(5.42) \\
\midrule
\multirow{3}{*}{\shortstack{LaTee \\ (\textit{Model Size})}} & opt-350M &  69.40(9.21) &  72.87(16.35)\\ 
& opt-1.3B & 66.40(9.52) & 34.50(5.23)\\
& opt-6.7B & 61.40(8.36) & 33.45(5.12)\\

\bottomrule
\end{tabular}
\end{sc}
}
\end{center}
% \vskip -0.2in
\end{table}

\begin{figure*}[!t]
    \centering
    \includegraphics[width=\textwidth]{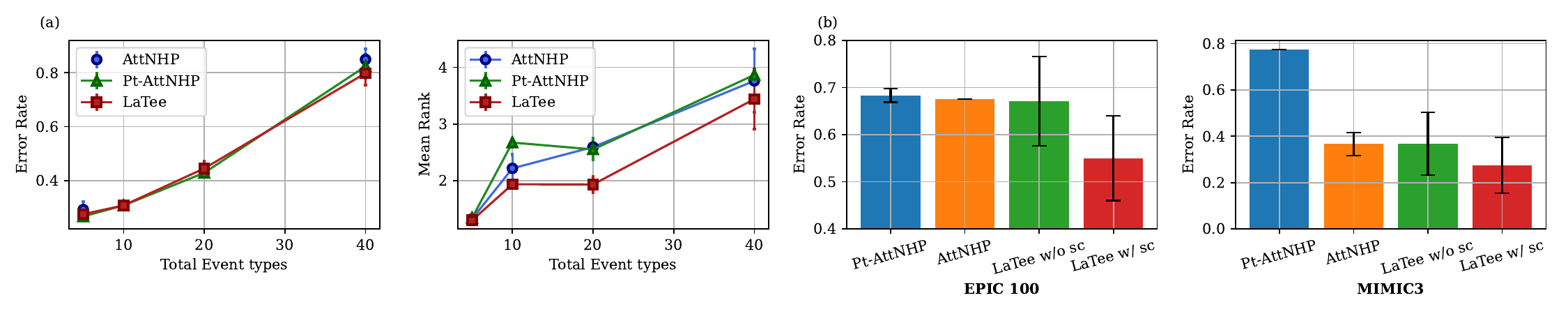}
    \vskip -0.2in
    \caption{(a) Illustration of Scalability on the number of event types on four synthetic datasets. (b) Illustration of the performance of using semantic and not using semantic information on two real-world datasets.}
    \label{fig:abalation}
\end{figure*}

\subsection{Results and Analysis}
The primary findings are summarized in Table \ref{tab:main_res}. Notably, only using the local LLM for event prediction $p_{LM}(Y|X)$ (0-shot CoT) yields the least effective results across all three datasets. Intriguingly, incorporating examples and adopting a tree-like reasoning structure (ToT) in the prompt do help enhance performance on the EPIC-100 and StackOverflow datasets to some extend. Furthermore, Supervised Fine-Tuning (SFT) exhibits similarly weak performance, while Proximal Policy Optimization (PPO) Fine-Tuning on the LLM shows marginal improvement but still lags behind attention-based Temporal Point Process (TPP) models. We hypothesize this underperformance is due to SFT's limited generalization capabilities and the shifted distribution mapping inherent in PPO's reward signals (refer to Analysis 2). It is noteworthy that Pt-AttNHP consistently falls short of AttNHP's performance across all datasets. This may be attributed to Pt-AttNHP's reliance on a prompt-like retrieval memory for time-horizon generalization, which potentially leads to overlooking individual-level characteristics in unseen event histories. Lastly, the proposed LaTee, fine-tuned with GFN objectives, matches AttNHP's accuracy on StackOverflow by focusing solely on latent structure learning. Remarkably, it surpasses AttNHP by a relative margin of 25\% on MIMIC-3 and 18\% on EPIC-100 containing semantic content (as detailed in Analysis 1).

\begin{table*}[ht]
\centering
\caption{Performance Evaluation for Alternate EM-loops Frequencies on Synthetic@5 (Earlystop was made at the fifth epoch).}
\label{tbl:em}
\begin{sc}
\begin{tabular}{cccc}
\hline
\textbf{} & \textbf{NLL $\downarrow$} & \textbf{ER (\%) $\downarrow$} & \textbf{MR $\downarrow$} \\ \hline
E-steps only (with groundtruth likelihood) & 1389.31 & 62.5 & 2.025 \\ 
EM-loops (Alternate Freq = 1) & 117.58 & 70.0 & 2.075 \\ 
EM-loops (Alternate Freq = 20) & 108.44 & 67.5 & 2.000 \\ 
EM-loops (Alternate Freq = 50) & 106.35 & 70.0 & 1.900 \\ \hline
\end{tabular}
\end{sc}
\end{table*}

\begin{table*}[ht]
\centering
\caption{Performance Evaluation for using different LMs for Inference (E-steps) and Generation (M-steps) on Synthetic@5 (Earlystop was made at the fifth epoch).}
\label{tbl:model_size}
\begin{sc}
\begin{tabular}{ccccc}
\hline
\textbf{E-steps LM (Fine-tuned)} & \textbf{M-steps LM (Frozen)} & \textbf{NLL $\downarrow$} & \textbf{ER (\%) $\downarrow$} & \textbf{MR $\downarrow$} \\ \hline
Opt-1.3b & Opt-1.3b & 128.64 & 97.5 & 2.23 \\
Opt-1.3b & Zephyr-3b & 149.25 & 87.5 & 2.50 \\ 
Opt-1.3b & Mistral-7B-Instruct & 117.62 & 70.0 & 2.08 \\ 
Zephyr-3b & Mistral-7B-Instruct & 116.21 & 70.0 & 1.95 \\ \hline
\end{tabular}
\end{sc}
\end{table*}

\textbf{Analysis 1: The Role of LLM in Enhancing Event Logic Discovery through Semantic Cognition.} From the data presented in Fig. ~\ref{fig:gpt_baseline}, it is evident that GPT enhances next events type prediction by substituting semantically meaningless numerical event IDs with meaningful event names. This study aims to explore whether the semantic content embedded in event history can bolster structure learning and, consequently, improve event prediction accuracy on a local deployable LLM. As shown in Fig. \ref{fig:abalation}(b), we observe a noteworthy reduction in error rate (approximately 25\%) for both EPIC-100 and MIMIC-3 datasets when employing semantic event names for reasoning and inference. This decrease is significantly more pronounced compared to the improvement seen when transitioning from attention-based TPP models to LaTee models that do not apply semantic information. Moreover, we illustrate two examples of semantic tree structures learned by LaTee in Appendix \ref{appsec:structure_examples}.

\textbf{Analysis 2: The Necessity of GFN Fine-Tuning in LLMs for Logic Tree Discovery and the Role of Prompts in Rule Discovery.} As indicated by the baselines in Table \ref{tab:main_res}, approaches such as zero-shot Chain-of-Thought (CoT) prompting, $k$-shot prompting, and Tree-of-Thoughts (ToT) prompting demonstrate limited efficacy in yielding meaningful results. Similarly, Supervised Fine-Tuning (SFT) and Proximal Policy Optimization (PPO) fine-tuning on Large Language Models (LLMs) for next events prediction are outperformed by attention-based Temporal Point Process (TPP) models. However, GFN fine-tuning, which focuses on teaching models how to reason rather than predict, enables LLMs to match and even exceed the prediction accuracy of attention-based TPP models, particularly when integrating semantic information. To understand this improvement, Fig. \ref{fig:diversity} offers a visualization of the diverse rule distributions generated by fine-tuned LLMs. We notice that rule distributions in both SFT and PPO fine-tuning are predominantly concentrated in five regions, whereas GFN fine-tuning exhibits a more diverse spread across the entire rule space.

\textbf{Analysis 3: The Scalability of the Proposed Method and the Impact of LLM and Symbolic Logic Tree Sizes on Performance.} This analysis explores the scalability of our proposed method by examining the effect of an increased number of event types across four synthetic datasets without any semantic information. As shown in Fig. \ref{fig:abalation}, LaTee demonstrates comparable scaling abilities in Error Rates and Mean Rank to those of attention-based TPP models. Notably, LaTee consistently achieves a lower Mean Rank, likely due to the additional confidence imparted by the learned structure information in making predictions. Additionally, we analyze the impact of varying tree sizes and LLM sizes. Assuming the predefined predicate space $\mathcal{Z}$ has a cardinality $|\mathcal{Z}|=N$, the maximum allowable depth and width of the logic tree are restricted to $d$ and $w$ ($w << N$), respectively, then the entirety of the search space can be approximated as $O(N^{{w}^{d}})$.  In Table \ref{tab:scal}, we restrict depth $d$ and width $w$ below $4$ and $7$ and the empirical findings suggest that increasing the tree widths has a more beneficial effect than increasing tree depth or model size on semantic event sequences. This could be attributed to the fact that ground-truth rules often consist of multiple short rules, and a wider tree is better equipped to encompass more semantically similar predicate events at the same level. It's also important to note that for non-semantic event sequences, enlarging the model size tends to be more advantageous than increasing tree sizes.

\textbf{Analysis 4: Ablating E-M Update Steps in LaTee.}
Unlike traditional EM algorithms where the E-step typically has a closed-form solution, E-step in GFlowNet-EM progressively moves closer to the target distribution $p$. This requires sufficient gradient steps in the `approximate E-step' to closely align the approximate distribution with the target while it also should regularly switch to M-steps for updating likelihood functions using the new sampled latent variables in E-steps. This non-stationary update thus gives us a challenge of scheduling E-M steps for a better convergence rate. 

Consequently, we added experiments comparing the convergence speed of both SubTB loss (E-steps) and NLL loss (M-steps) under varying frequencies of alternation. We provide the plot of convergence analysis for EM in the Appendix \ref{appsec:add_exper} Fig. \ref{appfig:em_m_step_loss} and \ref{appfig:em_subtb_loss} and report final performance in Table \ref{tbl:em}. Interestingly, we observe that more frequent alternations of E-M loops lead to a faster convergence of the SubTB loss (E-steps) but a slower rate for M-step. Additionally, the frequency of alternation appears to have minimal impact on the overall evaluation performance.

\textbf{Analysis 5: Ablating LLMs for E-M Steps.}
To investigate whether the world knowledge in the LM is most useful in the generation model (M-steps LM), the inference model (E-steps LM), or both, we compared the effects of using different sizes/versions of LMs for inference (E-steps) and generation (M-steps). In our experiment, we used Opt-1.3B as the base inference model (which has a minor language understanding ability on LM benchmark task), and used three different estimation (generation) models to make the event prediction, i.e., Opt-1.3B, Zephyr-3B, Mistral-7B-Instruct. The results are shown in Table \ref{tbl:model_size}.

Our evaluation strategy in Table \ref{tbl:model_size} focused exclusively on altering the model size to guarantee fairness in comparison. We observe that employing larger language models (LMs) for both inference (E-steps) and generation (M-steps) phases can enhance event prediction performance. Notably, an increase in the size of the LM used for generation (M-steps) exhibited a more pronounced positive impact compared to enlarging the LM for inference (E-steps). The results suggest that the extensive world knowledge encoded in larger LMs is more beneficial for generation tasks (M-steps). This finding encourages future improvements in reasoning abilities in the M-steps by calling API-based LLMs like GPT-4 and Claude-3 with an extracted logic tree from a fine-tuned local lightweight LLMs as the prompt.

% \textbf{Question 6: How does LLM-TPP generalize to unseen event sequence? What is the few-shot/zero-shot ability of LLM-TPP after fine-tuning by GFlowNet objectives?}
% One table for generalization
\section{Conclusion}
The incorporation of general knowledge from Large Language Models (LLMs) is key to deciphering complex structures in noisy event sequences. To facilitate this, we present LaTee, an amortized EM-style framework that leverages LLMs' prior knowledge for latent tree structure learning for event sequence explanation. We simplify the complex posterior with GFlowNets and perform inference based on the learned structure without further gradient updates. Empirical results show that this method notably enhances generalization in event histories with semantic information.

\newpage

\section*{Impact Statement}
This paper presents work whose goal is to advance the field of Machine Learning. There are many potential societal consequences of our work, none of which we feel must be specifically highlighted here.

\section*{Acknowledgement}
The authors thank the anonymous reviewers for their careful reading of our manuscript and their many insightful comments and suggestions. Shuang Li’s research was in part supported by the National Science and Technology Major Project under grant No. 2022ZD0116004, the NSFC under grant No. 62206236, Shenzhen Science and Technology Program under grant No. JCYJ20210324120011032,   Shenzhen Key Lab of Cross-Modal Cognitive Computing under grant No. ZDSYS20230626091302006, and Guangdong Key Lab of Mathematical Foundations for Artificial Intelligence. Zitao Song and Bo AN are supported by the National Research Foundation Singapore and DSO National Laboratories under the AI Singapore Programme (AISG Award No: AISG2-GC-2023-009).

\bibliography{example_paper}
\bibliographystyle{icml2024}

%%%%%%%%%%%%%%%%%%%%%%%%%%%%%%%%%%%%%%%%%%%%%%%%%%%%%%%%%%%%%%%%%%%%%%%%%%%%%%%
%%%%%%%%%%%%%%%%%%%%%%%%%%%%%%%%%%%%%%%%%%%%%%%%%%%%%%%%%%%%%%%%%%%%%%%%%%%%%%%
% APPENDIX
%%%%%%%%%%%%%%%%%%%%%%%%%%%%%%%%%%%%%%%%%%%%%%%%%%%%%%%%%%%%%%%%%%%%%%%%%%%%%%%
%%%%%%%%%%%%%%%%%%%%%%%%%%%%%%%%%%%%%%%%%%%%%%%%%%%%%%%%%%%%%%%%%%%%%%%%%%%%%%%
\newpage
\appendix
\onecolumn

\section{Learned Logic Tree Examples}
\label{appsec:structure_examples}

\begin{figure}[h]
    \centering
    \includegraphics[width=\textwidth]{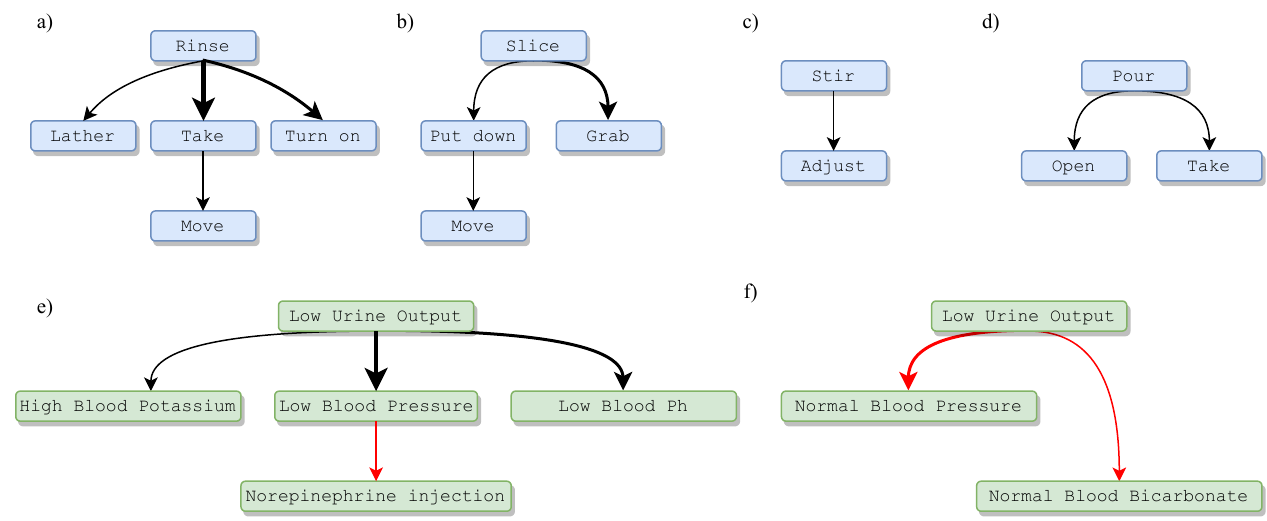}
    \caption{\textit{Illustration of learned symbolic logic tree structures from event histories on two real-world datasets containing semantic information.} Fig. (a)-(d) are learned from EPIC-100 and Fig. (e)-(f) are learned from MIMIC-3. We use the thickness of the edges to represent posterior probability and the color to represent the weights corresponding to each rule (Black color stands for activation and Red color stands for inhibition, \textit{i.e.}, low blood pressure will facilitate low urine while normal blood pressure suppresses low urine). }
    \label{fig:enter-label}
\end{figure}

\section{Broader Impact}

\textbf{Differentiable Extraction of Non-Linear Structures from LLMs}: Our approach extends the application of in-context learning in large language models (LLMs) beyond traditional posterior inference \citep{xie2021explanation} and independent demonstrations \citep{wang2024large}. We focus on non-linear prompt structures, enabling the extraction of complex entities like symbolic proof trees and latent positions of political figures. This differentiable method enhances the versatility of LLMs in handling diverse, non-linear structures.

\textbf{Advancing Neuro-Symbolic Inference with Foundation Models}: Foundational models, including Vision-and-Language Models (VLMs) and LLMs, serve as informative belief priors for world modeling and latent concept understanding. Our work augments posterior inference capabilities, moving past models like A-NESI \citep{van2023nesi} that rely on uninformative Dirichlet priors. This progression is pivotal for tackling more intricate, multimodal, and scalable neurosymbolic challenges.

\textbf{Enhancing Data Privacy in Event Sequence Explanation and Prediction}: By fine-tuning locally accessible, lightweight LLMs (under 7B parameters) while maintaining data privacy, our model offers wide applications in sensitive areas like healthcare and credit card fraud detection. The logic trees extracted from local LLMs can be integrated with public LLMs for prediction tasks. This aspect also paves the way for exploring improvements in reasoning abilities for API-based LLMs like GPT-4 and Claude-3 using these extracted logic trees.

These facets of our research not only contribute to the evolution of language model applications but also pave the way for new advancements in privacy-sensitive areas and neurosymbolic computing.

\section{More Related Works}
\textbf{Temporal Point Processes.} In recent decades, a diverse range of Neural Temporal Point Processes (TPPs) have been proposed to model event sequences with various properties. Many of these TPPs are based on a parametric intensity function that evolves through a series of latent states \citep{du2016recurrent,xiao2017modeling, boyd2020user,chen2020neural}. To effectively capture long-range dependencies within these sequences, the attention mechanism has been adapted for TPPs \citep{zuo2020transformer,yang2021transformer,mei2017neural}. Moreover, intensity-free TPP models have also shown promising results, particularly in the EasyTPP framework \citep{shchur2019intensity,xue2023easytpp}. However, the application of Large Language Models (LLMs) in learning event sequences remains largely unexplored. Recent research, such as LAMP \citep{shi2023language}, introduces a GPT-based abductive reasoning approach built upon attention-based TPP models for event prediction. This approach, however, necessitates additional textual data for event description and relies on costly API services. Our focus, instead, is on harnessing the reasoning capabilities of local LLMs for event prediction.  

\textbf{Non-Linear Reasoning in LLMs. } Recent research has focused on exploring complex, non-linear reasoning paths such as tree structures within Large Language Models (LLMs) \citep{zhu2022solving,xu2023no,yao2023tree,hao2023reasoning,xie2023decomposition}. Various methods, including beam search \citep{xie2023decomposition}, depth-/breadth-first search \citep{yao2023tree}, Monte Carlo Tree Search \citep{hao2023reasoning}, and MCTS with an enhanced value function \citep{feng2023alphazero}, have been implemented to navigate these tree structures effectively using LLMs' self-assessment capabilities to identify more effective reasoning pathways. Nonetheless, research on differentiable learning for non-linear reasoning within LLMs remains scarce. Recent studies, such as by  \citet{hu2023amortizing}, suggest fine-tuning LLMs using GFlowNets objectives to augment the diversity of reasoning chains. In our research, applying LLM-based tree search to discern the inherent structure in event sequences presents challenges due to the limited event data available for fine-tuning LLMs for specific event prediction tasks. Therefore, our focus shifts towards the development of differentiable logic trees to facilitate non-linear reasoning in LLMs, achieved by iteratively expanding and refining the logic tree structure.

\section{Limitations}
Resource constraints limited our experiments to models with up to 6.7B parameters and event sequences of a maximum of 40 events. This limited the capacity of input events because of the constraints on the maximum number of input tokens for a language model. However, we anticipate our findings to be applicable to larger models and longer sequences. Notably, optimizing larger models with limited data presents challenges, and exploring more complex latent problems is an ongoing challenge.

\section{ELBO Derivation}
\label{appsec:elbo}
Given data pair $(X, Y)$, we can represent write the joint likelihood $\log p(X, Y)$ as 
\begin{align}
    \log p(X, Y) &= \log \frac{p(X, Y, \mathcal{R})}{p(\mathcal{R} | X, Y)} \\
    &= \log \frac{p(X,Y,\mathcal{R}) q(\mathcal{R} | X,Y)}{p(\mathcal{R}|X,Y) q(\mathcal{R} | X, Y)} \\
    \sum_{R} q(\mathcal{R} | X, Y)  \log p(X, Y) &= \sum_{R} q(\mathcal{R} | X, Y) \log \frac{p(X,Y,\mathcal{R}) q(\mathcal{R} | X,Y)}{p(\mathcal{R}|X,Y) q(\mathcal{R} | X, Y)} \\
  \log p(X, Y)  &= \sum_{R} q(\mathcal{R} | X, Y) \log \frac{q(\mathcal{R} | X,Y)}{ p(\mathcal{R} | X, Y)} + \sum_{R} q(\mathcal{R} | X, Y) \log \frac{p(X,Y,\mathcal{R})}{ q(\mathcal{R} | X, Y)} \\
  &=D_{\text{KL}}(q || p) + \underbrace{\mathbb{E}_{\mathcal{R} \sim q(\mathcal{R} | X, Y)} [\log \frac{p(X, Y|\mathcal{R}) p(\mathcal{R})}{q(\mathcal{R} | X, Y)}]}_{\text{ELBO}} \\
  &\ge \mathbb{E}_{\mathcal{R} \sim q(\mathcal{R} | X, Y)} [\log \frac{p(X, Y|\mathcal{R}) p(\mathcal{R})}{q(\mathcal{R} | X, Y)}]
\end{align}
Thus, the ELBO $\mathcal{L}$ for the joint likelihood of $p(X, Y)$ is $\mathbb{E}_{\mathcal{R} \sim q(\mathcal{R} | X, Y)} [\log \frac{p(X, Y|\mathcal{R}) p(\mathcal{R})}{q(\mathcal{R} | X, Y)}]$.

\section{GFlowNets Learning Objective}
\label{appsec:subtb}
We learn the amortized sampler of posterior distribution $p(\mathcal{R}|X, Y)$ by a Sub-Trajectory Balance Objective \citep{madan2023learning} of GFlowNet. The original Sub Trajectory objective is given by:
\begin{align}
    \mathcal{L}_{SubTB}(\tau_{m:n}) &= \Bigg( \log \frac{F(s_{m};\theta)\Pi_{i=m}^{n-1}p_{F}(s_{i+1}|s_{i};\theta)}{F(s_{n};\theta) \Pi_{i=m}^{n-1} p_{B}(s_{i}|s_{i+1;\theta})} \Bigg)^{2} \\
    \mathcal{L}(\tau) &= \frac{\sum_{ 0 \le i <j \le n} \lambda^{j-i} L_{SubTB}(\tau_{i:j})}{\sum_{0 \le i <j \le n }\lambda^{j-i}}
\end{align}

In our case, we enforce $F(s_{n}; \theta) =R (s_{n})$ if $s_{n}$ is terminal, so we have $R(s_{n}^{\texttt{T}}) = F(s_{n}) p_{F}(\texttt{T} | s_{n})$. Since we are generating a tree structure level by level, thus the backward probability is one, \textit{i.e.,} $p_{B}(s|s') = 1$, and $\lambda=1$, we have
\begin{align}
 \mathcal{L}_{SubTB}(\mathcal{R}_{0:n}) &=\sum_{0 \le i < j \le n} \Bigg( 
 \log \frac{F(\mathcal{R}_{i};\theta) \Pi_{k=i+1}^{j}p_{F}(\mathcal{R}_{k} | \mathcal{R}_{k-1})}{F(\mathcal{R}_{j};\theta) \Pi_{k=i+1}^{j}p_{F}(\mathcal{R}_{k-1} | \mathcal{R}_{k})}\Bigg)^{2} \\
    &= \sum_{0 \le i < j \le n} \Bigg( \log \frac{R(\mathcal{R}_{i}^{\texttt{T}})\Pi_{k=i+1}^{j}q_{\theta}(\mathcal{R}_{k}|\mathcal{R}_{k-1})q_{\theta}(\texttt{T}|\mathcal{R}_{j})}{R(\mathcal{R}_{j}^{\texttt{T}})q_{\theta}(\texttt{T}|\mathcal{R}_{i})}\Bigg)^{2},
\end{align}
We train the GFlowNet with stochastic gradient
\begin{equation}
    \mathbb{E}_{\mathcal{R}_{0:n} \sim q_{\theta}}[\nabla_{\theta} \mathcal{L}_{SubTB}(\mathcal{R}_{0:n}) ]
\end{equation}

\section{Experimental Details}

\subsection{Dataset Details}

\label{appsec:dataset}

\begin{table}[h]
    \centering
    \begin{tabular}{lccc}
    \toprule
         & \# Target Predicates & \# Body Predicates & Events Average Length  \\ \midrule
    Synthetic@5 (w/o sc) & 2  & 3 & 30.19 \\
    Synthetic@10 (w/o sc)& 5 & 5 & 30.34 \\
    Synthetic@20 (w/o sc) & 7 & 13  & 30.29 \\
    Synthetic@40 (w/o sc) & 8 & 32  & 30.82\\
    \midrule
     StackOverflow (w/o sc)& 10 & 22 & 40.00 \\
    EPIC-KITCHEN-100 (w/ sc)& 7 & 60 & 36.76 \\
    MIMIC3 (w/ sc)& 3 & 62 & 20.01 \\
    \bottomrule
    \end{tabular}
    \caption{Event Dataset Statistics}
    \label{tab:data}
\end{table}

We evaluate our methods on one synthetic dataset and three user behavior datasets. We consider each event type presented in the event history as a unique predicate and emphasize on the model's ability to predict only pertinent target predicates. The overall data statistics is presented in Table \ref{tab:data}. We provide details on the preparation and utilization of each below.

\textbf{Synthetic Dataset.} This dataset comprises four sets of synthetic event history data generated using the Temporal Logic Point Process \citep{li2020temporal}. Specifically, we employ pre-defined logical rules along with their weights, as outlined in Eq. (\ref{eq:intensity}), to construct the intensity function, and then apply thinning algorithms to generate new events. To evaluate the scalability of the proposed model, we have created four distinct groups of synthetic data, with the number of event types varying from five to forty, and an average sequence length of 30 events.

\textbf{StackOverflow} \citep{leskovec2014snap}. This dataset encompasses two years of user awards from a question-and-answer website, documenting each user's sequence of badges. There are 22 distinct types of badges in total. However, since each event type is represented solely by a numerical ID, the dataset lacks semantically meaningful information. We focus on a subset of 142 records, each with an average sequence length of 40 event tokens.

\begin{figure}
    \centering
    \includegraphics[width=\textwidth]{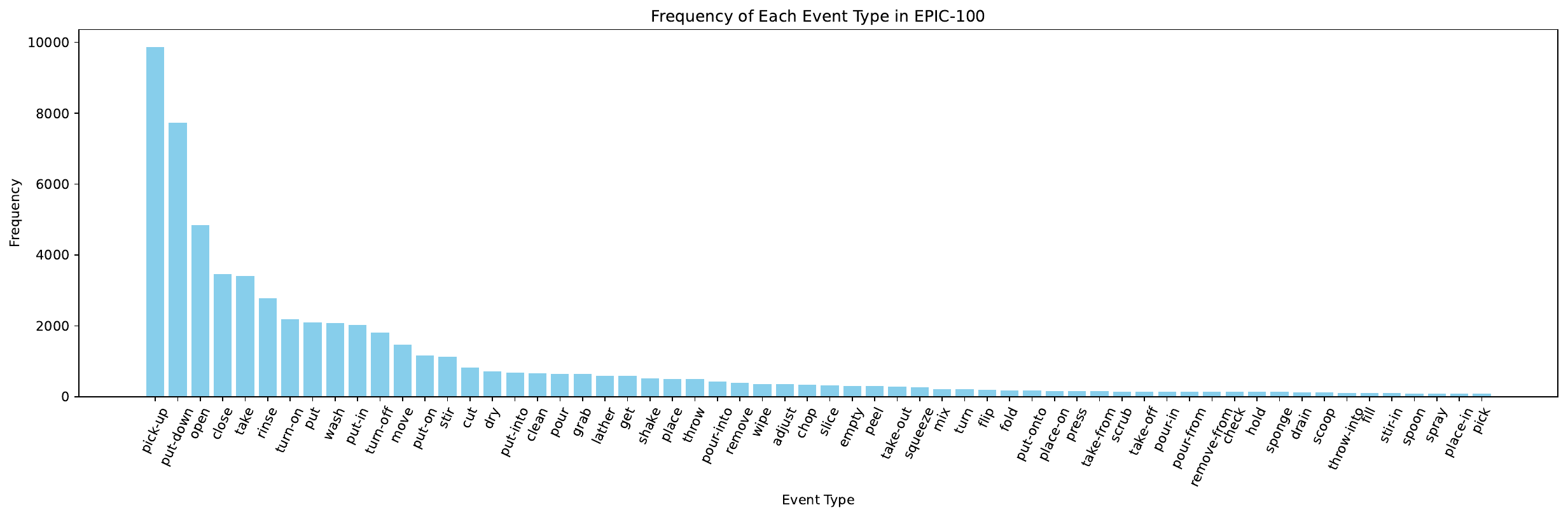}
    \caption{Distribution of Semantic Event types in \textbf{EPIC-100}}
    \label{fig:epic100_dist}
\end{figure}

\textbf{EPIC-KITCHEN-100} \citep{Damen2021PAMI}. This dataset originates from a large-scale, first-person (egocentric) vision dataset, featuring multi-faceted, audio-visual, non-scripted recordings in natural settings, specifically the wearers' homes. It captures daily kitchen activities over multiple days. We have utilized the annotated action sequences, focusing only on text, and extracted them to create a temporal event history of cooking verbs. This was achieved by omitting the entities that the human subjects interacted with. The frequencies of each verb, derived from the Epic-100 dataset, are visualized in Fig. \ref{fig:epic100_dist}. In this dataset, we specifically focus on eight verbs: \texttt{put-in}, \texttt{rinse}, \texttt{put-on}, \texttt{pour}, \texttt{stir}, \texttt{peel}, \texttt{chop}, and \texttt{slice}, as our target predicates. The model is tasked with reasoning about the actions preceding each target verb and learning the underlying structure that culminates in these targets. We concentrated on a subset of 400 event histories, each with an average sequence length of 36.76 events, resulting in 60 distinct event types in total.

\begin{figure}
    \centering
    \includegraphics[width=\textwidth]{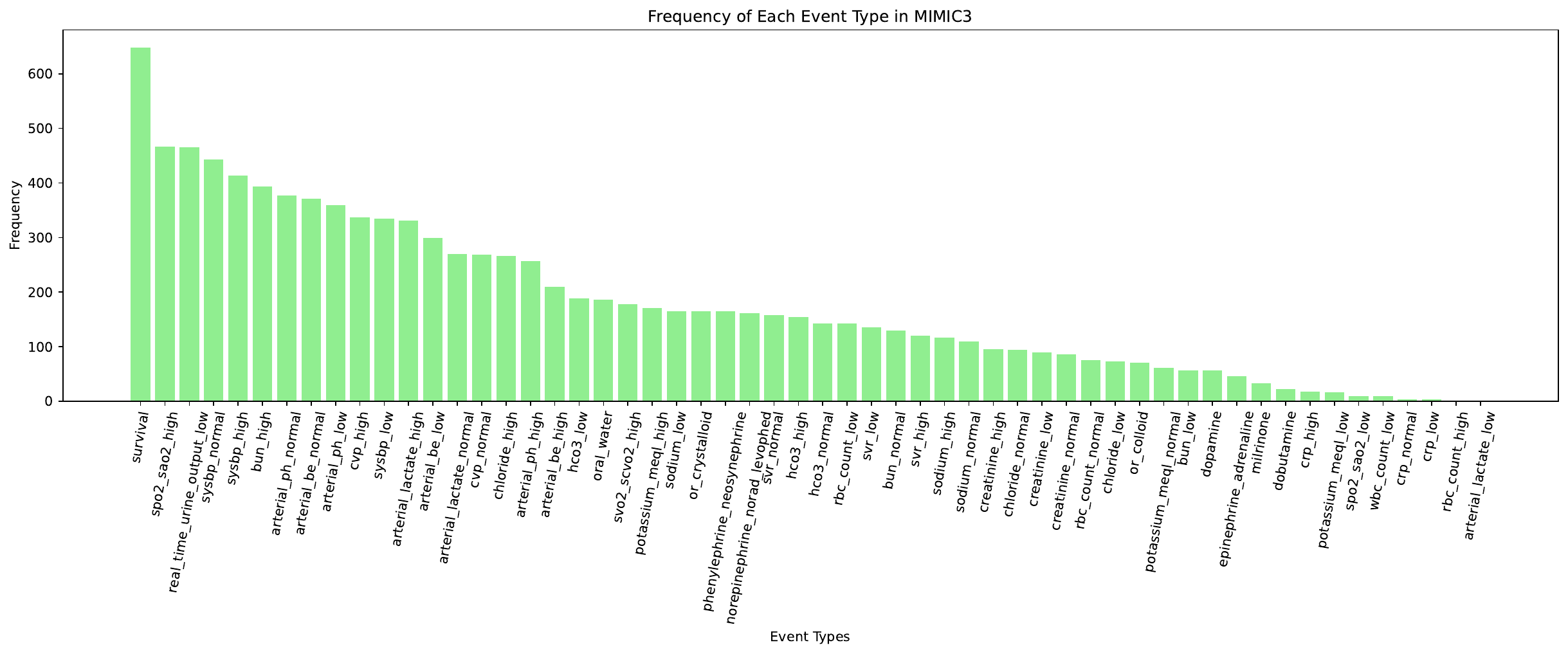}
    \caption{Distribution of Semantic Event types in \textbf{MIMIC-3}}
    \label{fig:mimic_dist}
\end{figure}

\textbf{MIMIC-3} \citep{johnson2016mimic}. This dataset comprises electronic health records of patients admitted to the intensive care unit (ICU). We specifically focus on patients diagnosed with sepsis, extracting medications, lab tests, outputs, and diagnoses to form text-based temporal event histories. The frequencies of the various event types related to sepsis are illustrated in \ref{fig:mimic_dist}. In this dataset, we concentrate on three key event types: \texttt{survival}, \texttt{urine\_output\_low}, and \texttt{normal\_blood\_pressure}. Our analysis is based on a subset of 477 event histories, each with an average sequence length of 20 event tokens, resulting in a total of 62 unique event types.

\subsection{Praising LLM Outputs}
We detail two distinct methodologies for parsing outputs from large language models (LLMs) and querying corresponding probabilities. 
\begin{enumerate}
    \item  For LLMs that are locally accessible, such as OPT series, our approach aligns with that of \citep{hu2023amortizing}. Here, we directly query the probability of subsequent tokens given a target sentence, avoiding the need for parsing. To illustrate, consider an 'action space' defined by $\{A, B, C, D\}$. To get the probability of action $A$, we tokenize it into k tokens $[w^{A}_{1}, w^{A}_{2},...,w^{A}_{k}]$. Then $p(A)$ is computed as:

$$
p(A) = p(w^{A}_{k}|w^{A}_{k-1},...,w^{A}_{1})p(w^{A}_{k-1}|w^{A}_{k-2},...,w^{A}_{1})...p(w^{A}_{1}).
$$

\item For LLMs that are not locally accessible, like GPT, we employ a different technique. Regular Expressions are used to isolate target sentences. Then, we leverage the 'logprobs' parameter in the OpenAI Chat Completions API to ascertain the probabilities of target tokens. For example, a common pattern in our analysis is '\#Event {NAME}\#', which allows us to capture the output event by extracting the NAME component. In cases where the NAME fails to be parsed, we adopt the approach from logic-LM \citep{pan2023logic}, making a random guess across all possible event types. 
\end{enumerate}

\subsection{Additional Experiments}
\label{appsec:add_exper}

\begin{figure}[h]
    \centering
    \includegraphics[width=0.5\textwidth]{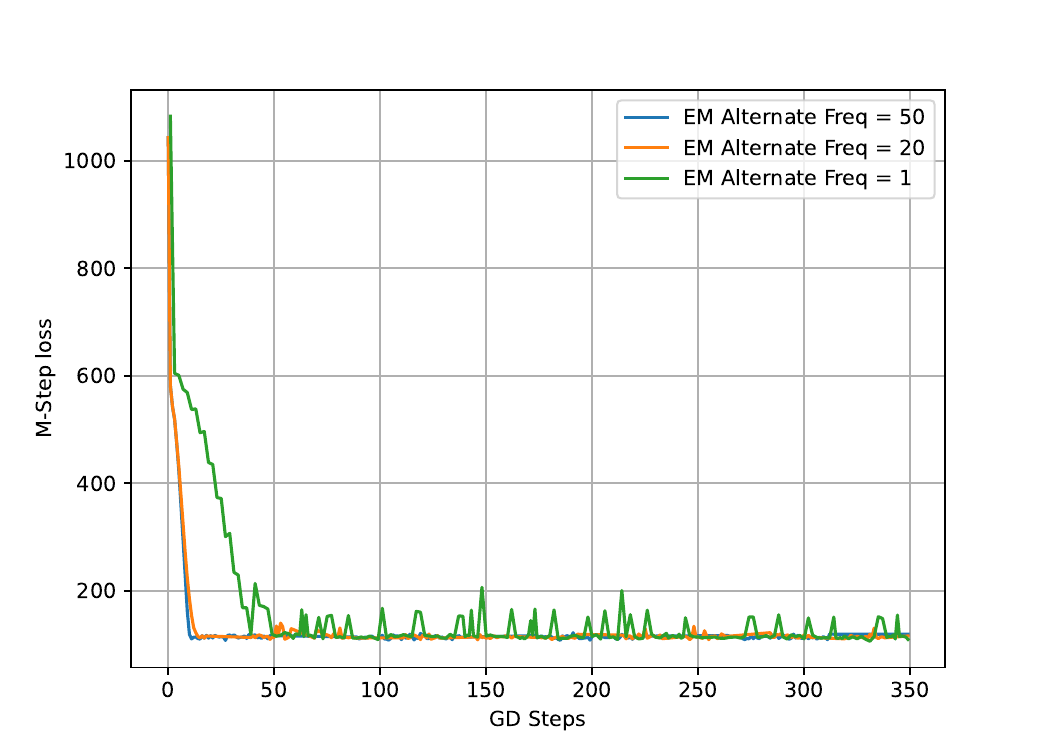}
    \caption{M-Steps converges rate under different EM-loops alternating frequencies}
    \label{appfig:em_m_step_loss}
\end{figure}

\vspace{1pt}

\begin{figure}[h]
    \centering
    \includegraphics[width=\textwidth]{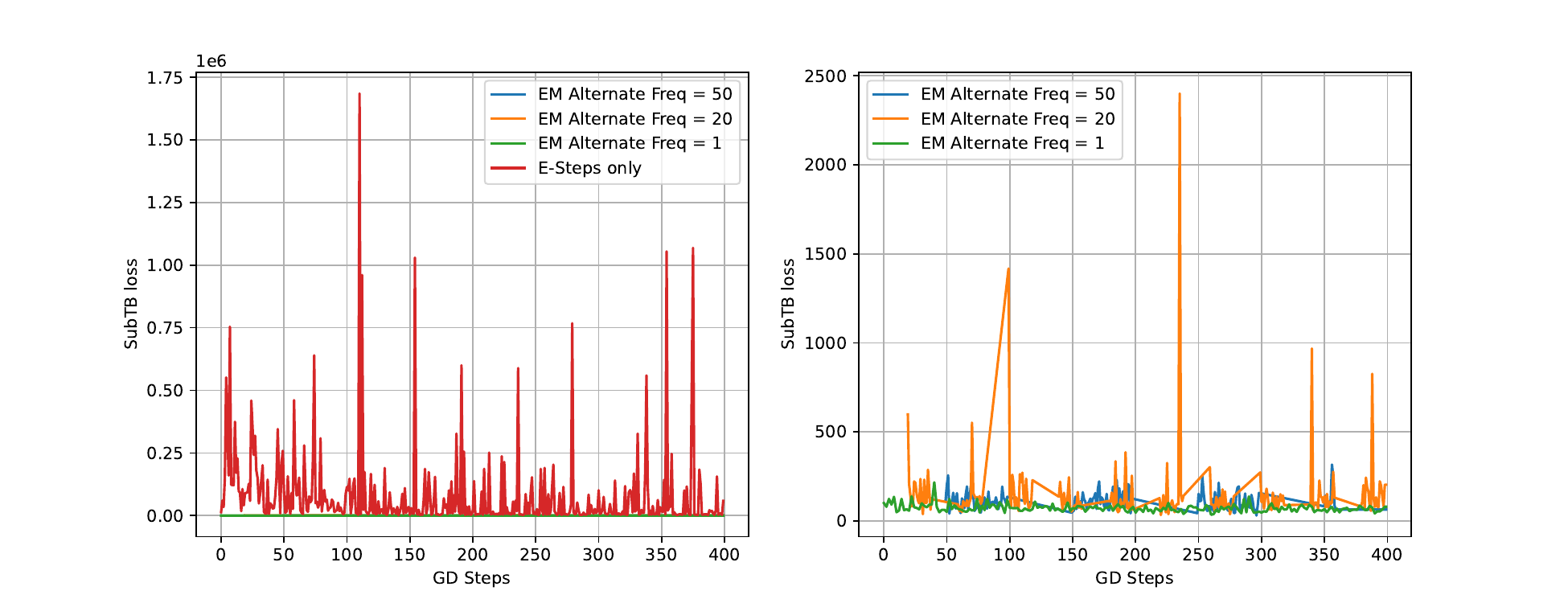}
    \vspace{-5pt}
    \caption{E-Steps converge rate under different EM-loops alternating frequencies. \textbf{Left}: Four different alternating frequencies including E-steps only. \textbf{Right}: Excluding E-steps only from Left.}
    \label{appfig:em_subtb_loss}
\end{figure}

% \begin{table}[h]
% \centering
% \caption{Performance Evaluation for using different LMs for Inference (E-steps) and Generation (M-steps)}
% \label{apptbl:model_size}
% \begin{tabular}{ccccc}
% \hline
% \textbf{Inference Model (Fine-tuned)} & \textbf{Generation Model (Frozen)} & \textbf{NLL $\downarrow$} & \textbf{ER (\%) $\downarrow$} & \textbf{MR $\downarrow$} \\ \hline
% Opt-1.3b & Opt-1.3b & 128.64 & 97.5 & 2.23 \\
% Opt-1.3b & Zephyr-3b & 149.25 & 87.5 & 2.50 \\ 
% Opt-1.3b & Mistral-7b & 117.62 & 70.0 & 2.08 \\ 
% Zephyr-3b & Mistral-7b & 116.21 & 70.0 & 1.95 \\ \hline
% \end{tabular}
% \end{table}

\subsection{Training Details}
\label{appsec:ft_llm}
\textbf{Implementation Details.} All models are implemented using the PyTorch framework. All the experiments were conducted on a server with 512G RAM, two 64 logical cores CPUS (AMD Ryzen Threadripper PRO 5995WX 64-Cores), and four NVIDIA RTX A6000 GPUs with 50G memory.

\textbf{Hyperparameters Selection.}
We present the selected hyperparameters on synthetic datasets and three real-world datasets in Table \ref{tab:hyper_1} and Table \ref{tab:hyper_2} respectively.

\textbf{Fine-tuning Quantized Large Language Model.}

In our experiment, we implement QLoRA \citep{dettmers2023qlora} to fine-tune the Language Model, effectively reducing memory requirements during LLM finetuning without compromising on performance, as compared to the conventional 16-bit model finetuning process. Specifically, QLoRA employs 4-bit quantization to condense a pre-existing language model. This model's parameters are then set as unchangeable, and a limited set of modifiable parameters are incorporated via Low-Rank Adapters. During the finetuning phase, QLoRA directs gradient updates through these unmodifiable 4-bit quantized pre-trained language model parameters to the Low-Rank Adapters. Only the LoRA layers are adjusted during the training process.

\textbf{Prompt Details.}
We present the prompts utilized for reasoning, denoted as $P(\mathcal{R} | X, Y)$, and inference, represented by $P(Y | X, \mathcal{R})$, in Tables \ref{tab:prompt_structure} and \ref{tab:prompt_inference}, respectively. We iteratively grow the logic tree by applying the structure learning prompt to the successive node within the current tree. Importantly, our prompts are crafted using a simple, predefined template. In this template, \texttt{events\_history} represents an in-text version of the observed event sequence $X$, and \texttt{target\_event} corresponds to the event id/name associated with $Y$. We also provide three inference examples in Table \ref{tab:prompt_inference}.

\begin{table}[ht]
    \centering
     \caption{Decriptions and values of hyperparameters used for models trained on the four synthetic datasets.}
    \begin{tabular}{ccccc}
    \toprule
    HYPERPARAMETERS & \multicolumn{4}{c}{VALUE USED } \\
    \midrule
    & SYNTHETIC@5 & SYNTHETIC@10 & SYNTHETIC@20 & SYNTHETIC@40 \\ \midrule
    EPOCHS & 10 & 10 & 10 & 10 \\
    ALTERNATE EVERY & 1  & 1 & 1 & 1\\
    BATCH SIZE & 8 & 8 & 8 & 8\\
    LLM LR & 5e-4 & 5e-4 & 5e-4 & 5e-4 \\
    LLM SIZE (E-STEP) & opt-1.3b & opt-1.3b & opt-1.3b & opt-1.3b \\
    LLM SIZE (M-STEP) & zephyr-3b & zephyr-3b & zephyr-3b & zephyr-3b \\
    LOGIC MODEL UPDATE STEPS & 1 & 1 &  1 & 1\\
    LOGIC MODEL LR & 0.001 & 0.001 & 0.001& 0.001 \\
    LOGIC TREE DEPTH & 3 & 3 & 3 & 3\\
    LOGIC TREE WIDTH & 5 & 8 & 5 & 3\\
    TOP K & 2 & 2 & 2 & 2\\
    WRAMUP LEARNING RATE & True & True & True & True \\
    LoRA RANK & 512 & 512 & 512 & 512 \\
    LoRA SCALING FACTOR & 512 & 512 & 512 & 512 \\
    LoRA DROPOUT & 0. & 0. & 0. & 0. \\
    \bottomrule
    \end{tabular}

    \label{tab:hyper_1}
\end{table}

\begin{table}[ht]
    \centering
    \caption{Decriptions and values of hyperparameters used for models trained on the three real-world datasets.}
    \begin{tabular}{cccc}
    \toprule
    HYPERPARAMETERS & \multicolumn{3}{c}{VALUE USED } \\
    \midrule
    & EPIC-100 & STACKOVERFLOW & MIMIC-3 \\ \midrule
    EPOCHS & 20 & 20 & 20 \\
    ALTERNATE EVERY & 1  & 1 & 1 \\
    BATCH SIZE & 2 & 2 & 2\\
    LLM LR & 5e-4 & 5e-4 & 5e-4 \\
    LLM SIZE (E-STEP) & opt-1.3b & opt-1.3b & opt-1.3b \\
    LLM SIZE (M-STEP) & mistral-7b & zephyr-3b & zephyr-3b \\
    LOGIC MODEL UPDATE STEPS & 1 & 1 &  1\\
    LOGIC MODEL LR & 0.001 & 0.001 & 0.001 \\
    TREE DEPTH & 3 & 3 & 3\\
    TREE WIDTH & 4 & 3 & 2\\
    TOP K & 2 & 3 & 2\\
    WRAMUP LEARNING RATE & True & True & True \\
    LoRA RANK & 512 & 512 & 512  \\
    LoRA SCALING FACTOR & 512 & 512 & 512  \\
    LoRA DROPOUT & 0. & 0. & 0.  \\
    \bottomrule
    \end{tabular}
    \label{tab:hyper_2}
\end{table}

\begin{table}[h]
    \centering
    \begin{tabular}{cp{10cm}}
    \toprule
      &   \textbf{Bayesion Structure Learning} $P(\mathcal{R} | X, Y)$ \\ \midrule
     \textbf{Template}  &  
     I want you to do the reasoning over social events.
Given event list: \{\texttt{total\_events}\} \newline
\newline
We have the observations: \newline
\{\texttt{events\_history}\} \newline
\newline
If the activation time of one event happens before Event \{\texttt{target\_event}\}, it means that event could have caused Event \{\texttt{target\_event}\} to be activated. \newline 
If the activation time of one event do not happens before Event \{\texttt{target\_event}\}, it means that event cannot cause the other event to be activated. \newline  
Using this logic and based on the previous observation, You need to reason all possible events from above that can cause Event \{\texttt{target\_event}\} to be activated. \newline
Start your answer from the most confident one and stop if you cannot find any other events.

Answer: Event 
\\
\bottomrule
\end{tabular}
\caption{Prompts used for structure learning}
    \label{tab:prompt_structure}
\end{table}

\begin{table}[h]
    \centering
    \caption{Prompts Used for Next Event Inference. \texttt{rationales} store the text representation of the logic tree by going over all the paths. The reasoning path is highlighted in red color.}
    \begin{tabular}{cp{7.1cm}p{7.1cm}}
    \toprule

    & \textbf{Direct Inference} $P(Y|X)$ & \textbf{Reasoning based Inference} $P(Y|X, \mathcal{R})$  \\ \midrule
     \textbf{Template}  &  
     I want you to perform inference over social events. \newline
     \{\texttt{examples}\} \newline
     Now you have event: \{\texttt{total\_events}\} \newline
We have the observations: 
\{\texttt{events\_history}\} \newline
then, the most likely event (chosen from event list : \{\texttt{possible\_events}\}) to happen after \{\texttt{time}\} is Event:
     & 
     I want you to perform inference over social events. \newline
     \{\texttt{examples}\} \newline
     Now you have event: \{\texttt{total\_events}\} \newline
     and rules:
    \{\texttt{rationales}\} \newline
We have the observations: 
\{\texttt{events\_history}\} \newline
then, the most likely event (chosen from event list : \{\texttt{possible\_events}\}) to happen after \{\texttt{time}\} is Event:\\ \midrule 
\textbf{Example 1} 

& Given Events 0, 1 \newline
We have the observations: \newline
1. Event 0 is activated at time 0.4 \newline 
\newline
then, the most likely event (choose from event list: 0, 1) to happen after 0.4 is Event 1

&  Given Events 0, 1 and rules: \newline
{\color{pink}1. Event 1 $\leftarrow$ (Event 0) and (Time of Event 1 after Time of Event 0)\newline}
\newline
We have the observations: \newline
1. Event 0 is activated at time 0.4 \newline 
\newline
then, the most likely event (choose from event list: 0, 1) to happen after 0.4 is Event 1
\\ \midrule 
\textbf{Example 2}
& Given Events 0, 1, 2 \newline

We have the observations: \newline
1. Event 1 is activated at time 0.2 \newline
\newline 
then, the most likely event (chosen from event list : 0, 1, 2) to happen after 0.2 is Event 0
& Given Events 0, 1, 2 and rules: \newline
\newline
{\color{pink}1. Event 0 $\leftarrow$ (Event 1) and (Time of Event 0 after Time of Event 1), \newline 
2. Event 0 $\leftarrow$ (Event 2) and (Time of Event 0 after Time of Event 2) \newline}

We have the observations: \newline
1. Event 1 is activated at time 0.2 \newline

then, the most likely event (chosen from event list : 0, 1, 2) to happen after 0.2 is Event 0 
\\ \midrule
\textbf{Example 3}
 & Given Events 0, 1, 2 \newline
\newline
We have the following observation: \newline
1. Event 0 is activated at time 0.2, 0.3, 0.5 \newline 
2. Event 1 is activated at time 0.5, 0.6 \newline
3. Event 2 is activated at time 0.1, 0.4 \newline
\newline
then, the most likely event (chosen from event list : 0, 1, 2) to happen after 0.8 is Event 2
 & Given Events 0, 1, 2 and rules: \newline
\newline 
{\color{pink} 1. Event 2 $\leftarrow$ (Event 1) and (Event 0) and (Time of Event 2 after Time of Event 1) and (Time of Event 1 after Event 0)} \newline
\newline 
We have the following observation: \newline
1. Event 0 is activated at time 0.2, 0.3, 0.5 \newline
2. Event 1 is activated at time 0.5, 0.6 \newline 
3. Event 2 is activated at time 0.1, 0.4 \newline
\newline
then, the most likely event (chosen from event list : 0, 1, 2) to happen after 0.8 is Event 2
 \\
\bottomrule 
    \end{tabular}
    \label{tab:prompt_inference}
\end{table}
%%%%%%%%%%%%%%%%%%%%%%%%%%%%%%%%%%%%%%%%%%%%%%%%%%%%%%%%%%%%%%%%%%%%%%%%%%%%%%%
%%%%%%%%%%%%%%%%%%%%%%%%%%%%%%%%%%%%%%%%%%%%%%%%%%%%%%%%%%%%%%%%%%%%%%%%%%%%%%%

\end{document}